# Pathways to AGI

Gordon Fletcher* and Saomai Vu Khan

g.fletcher@salford.ac.uk

Salford Business School

University of Salford, UK, M5 4WT

# 1 Research Questions

Our focus are five related questions that stem from a critical software studies perspective. Underpinning this view is the acknowledged need to avoid assumptions regarding the inevitability of the current situation relating to AI. What we need to see is the closeness of the linkage between current commercial AI development and our prevailing social, political and economic circumstances. This does mean that the perspectives presented here are done so critically and conditionally. Most importantly, Artificial General Intelligence (AGI) is seen as being problematic both conceptually and definitionally. This conditioning of any view regarding AGI does lead the discussion in specific directions and to certain conclusions regarding the future. However, adopting this perspective enables the work to offer some final recommendations.

We set out to ask the following questions,

1. What are the critical pathways that produced the current dominant generative AI tools (capabilities, product forms, adoption patterns)?
2. Which decision points acted as leverage nodes (small changes that had large downstream effects), and which dead ends reveal alternative possibilities that did not become dominant?
3. How do pathways differ across three foundational-model trajectories such as the frontier proprietary models, open-weight models or specific domain and sovereign models?
4. Which alternative projects branched from key leverage nodes, what is their current state, and why did some succeed, stall, fail or become absorbed?
5. Based on this analysis, what socio-technical development programmes could plausibly move toward AGI-adjacent capability while meeting requirements for transparency, moderation, wellbeing and sustainable business models?

The perspectives of these questions also requires that generative AI is seen as a software artefact. The current dominant models are not neutral objects that simply reveal technical progress. They are historically situated artefacts that embody interface choices, training assumptions, business models, infrastructural dependencies and political priorities. In that sense, we are also asking what kinds of software these systems already are, what (and whose) worlds they are encoding and what futures their current forms will make easier or harder to realise.

# 2 Definition of Artificial General Intelligence (AGI)

There is no agreed definition for Artificial General Intelligence (Eliot, 2025; Tolić, 2026). Some commentators even deny the possibility of ever reaching AGI (Dettmers, 2025). But with this dismissal itself there is an implicit assumption that a specific given definition could be possible. There is also speculation, including from Anthropic's President Daniela Amodei, that some of the current commercial frontier foundation models have already reached a kind of AGI. She notes that her company's own large language model (LLM), Claude, has coding capabilities that are comparable to the capabilities of the top software engineers employed by Anthropic (Werner, 2026). Many of the definitions offered for AGI also come from these same commercial organisations which makes a consistent perspective even less likely as the feature set and most advanced capabilities of any single model become so tightly enmeshed in the definition that the company then offers.

## 2a Existing Vendor Definitions and human-centricity

The concept of Artificial General Intelligence (AGI) is contentious, in part, because all of the leading AI vendors have promoted their own interpretations of AGI. Microsoft acknowledges that "AGI is often seen as the point at which an AI can match human performance at all tasks" (Suleyman, 2025), whereas OpenAI defines AGI as "autonomous systems that outperform humans at most economically valuable work" (OpenAI, 2018). For Google, AGI is "AI that's at least as capable as humans at most cognitive tasks" (Google DeepMind, 2025). All three examples emphasis the human-centric perspective as the comparative gauge to measure AGI. Yet humans themselves are not homogenous in their intelligence or capabilities with an enormous range in any given sample of multiple individuals. Each of these definitions could be variously interpreted as the comparison against the "best" human for any given task, a randomly selected human or an "average" drawn from all humanity. In some discussions, the implied comparison is one that is being made against the capacity of multiple humans or even the entirety of humanity.

The lack of a single unambiguous definition means the pathway to AGI could differ depending on whether the AI must match human capability, or exceed it, and whether the comparison to be made is against all tasks, most or just some. In some ways there is a benchmark of defining what "is not" AGI. For the single task of knowledge recall, Google's search engine has exceeded the specific knowledge recall capability of any single human for at least a decade. But the search engine is not AGI and it has never been described as an example of AGI. Other examples, such as specific software including Excel's ability to instantly calculate complex financial equations, present the case for a tool exceeding the capabilities of one human for a single task.

The question is further complicated by the subjectivity of these requirements. For example, how is human capability measured for any given task and by what standards? If capability is not required across all tasks, then which ones do qualify for AGI? Anthropic co-founder and former OpenAI engineer Ben Mann prefers the term "transformative AI" over AGI, which can be measured according to the "Economic Turing Test". If an AI agent performs a job and is ultimately hired, without revealing itself as AI, then the agent has passed the Economic Turing Test. Once AI can pass this test for 50% of "money-weighted" jobs, then transformative AI is achieved (Mann, 2025). This is an observation that loops back to the OpenAI definition of AGI – where "economically valuable" activity is the basis of the definition with an even longer heritage back to underlying economic concepts of value and even Marx's Labour Theory of Value (Shaikh, 1998). In this case, the relationship of AI to the observation that "human labour is the sole source of new economic value, while machines and raw materials only transfer their existing value to products" becomes a key defining criteria for the possibility or otherwise of AGI. If AI (or AGI) does produce new economic value, then the question of its possibility becomes a significant existential one.

Leaked documents reveal "Microsoft and OpenAI's $100 billion definition of AGI" (Towers-Clark, 2025). The requirement for ChatGPT to generate $100 billion in profits before AGI can be declared further exposes an economic dimension in the pathway to AGI. Multiple circumstances underlie this definition, not least Microsoft's investment in OpenAI that is linked to an agreement that Microsoft will not be able to benefit from any AGI outcomes of OpenAI's work. But even with a discussion of money, the definition is still hazy. How will AGI "generate" the profit, and does this require evidence of the creation of new economic value by the AI itself? If income and costs attributed to a product are the basis for a simple determination, then it is only a time-based period for Microsoft to benefit from OpenAI developments. Based on Copilot revenues (Visual Capitalist, 2024), this may just be an estimated 10-year window.

The profit-based threshold is also a reminder that AGI can function as a legal and contractual mechanism. In the language of Microsoft–OpenAI partnership, the declaration of AGI is a claim subject to verification by independent experts. Even when the technical reality remains unchanged, the presence of a threshold reshapes the incentives for both organisations. Its presence may really reward postponement, increased ambiguity around capabilities or definitional drift.

Recent filings between Elon Musk and OpenAI also made definitions of AGI the basis for court rulings. The claim is that the GPT-4, 4T, and 4o models have already achieved a form of AGI (Elon Musk v. Altman, 2024b, para. 344). The commentary around this claim also points to the subsequent development of the private GPT-4b micro model as the basis for the business model of Retro Bioscience, a biotech startup backed by Sam Altman (OpenAI, 2025; Regalado, 2023). All this speculation points to a current reality that commercial models are hovering on the edge of some early form of AGI capability while facing the economic realities of managing commercial products and multi-billion-pound companies.

Vendor definitions are also very clearly political instruments. The tone and boldness of their statements direct the reader's attention toward certain capabilities, legitimises their current business models and helps to set out the conditions under which partners, regulators or investors might demand proofs for the capabilities or safety of their models. OpenAI's Charter definition directs attention to "economically valuable work" as the key basis for comparison, whereas Google DeepMind's definition emphasises "most cognitive tasks" and Microsoft's positioning clearly treats AGI as a point on a journey to somewhere else. In practice, the choice of framing shapes what gets measured by observers, what is prioritised by the vendor and what challenges or issues can be deferred.

## 2b Research based definitions of AGI

Research definitions of AGI are more varied than vendor definitions and they generally differ in one important respect. They all set out what should be measured, which failures matter and what kinds of systems are worth considering and building. As Wang (2019) argues in relation to AI more broadly, definitions function as the "cornerstone of a research paradigm". This is significant because disagreements over AGI definitions are based on the type of intelligence that is being sought and the pathway by which the intelligence could be achieved.

One influential research tradition defines intelligence in a broad and deliberately non-anthropocentric way. Legg and Hutter's "universal intelligence" treats intelligence as an agent's ability to achieve goals across a wide range of environments (Legg & Hutter, 2007). The attraction of this approach is that it does not tie intelligence too tightly to any form of human likeness, embodiment or any single task domain. It also provides a rare attempt at formal precision in a field that relies heavily on the use of metaphor. Yet this high-level abstraction is also a limitation. A highly general formal agent is not necessarily a socially or institutionally viable one. The academic definition is powerful for theory building, but it leaves untouched (possibly purposefully so) the questions of governance, legitimacy, resource constraints or the conditions under which the intelligence would have to operate in the lived world.

A second research tradition positions the question of adaptation much more centrally. Wang (2019) defines intelligence as "the capacity of an information-processing system to adapt to its environment while operating with insufficient knowledge and resources". Xu (2024) develops a closely related position and defines AGI as a computer that is adaptive to an open environment with limited computational resources and that satisfies certain principles. This can offer a useful shift because it distinguishes intelligence from (massively) stored competence or computationally heavy brute-force problem solving. A calculator, search engine or expert system may outperform humans in a bounded area, but this does not by itself show general intelligence. What matters in these research definitions is the ability for the intelligent system to continue learning, re-orienting and coping when the surrounding environment changes, the problem being addressed is never fully specified but resources are always finite.

Chollet's account sharpens this argument further by proposing intelligence as a form of skill-acquisition efficiency rather than raw skill (Chollet, 2019). His objection is that task performance can be "bought" through the use of large quantities of previously obtained, bespoke data or narrow optimisation, while still revealing relatively little about the actual capacity of a system to generalise. From this perspective, AGI would not be evidenced solely by solving many tasks, but by acquiring and transferring competence under novel conditions with limited prior preparation. This is particularly relevant to current claims made around commercial frontier models. Strong benchmark performance may indicate genuine progress, but it may also reflect extensive optimisation around known tasks rather than robust capability for generality in open-world settings.

More recent academic attempts to operationalise AGI retain all of this caution. Morris et al. (2023) propose "levels of AGI" based on both performance and generality, while explicitly emphasising cognitive and metacognitive tasks, ecological validity and the importance of treating AGI as a path rather than a single binary threshold to be exceeded. These are useful observations because they separate narrow superhuman systems from more genuinely general systems and recognises that open-ended, interactive and socially meaningful tasks may matter more than neat benchmark suites. However, it still leaves unresolved the central problem of which tasks should count, how generality is to be sampled from these systems and whether human-relative benchmarking can ever fully capture the plurality of intelligence in “real world” practice.

Taken together, research-based definitions do not remove ambiguity, but they do narrow the field of possible meanings. Across otherwise different traditions there is a repeated emphasis on adaptation, learning, transfer, open environments and action under constraint. What counts for less, in the research literature at least, is a single impressive demonstration, a profitable mass-market product or the automation of a narrow band of economically valuable tasks. For this report, understanding this distinction does have consequence. It suggests that AGI is better treated as a contested research space regarding the generality and adaptive capacity of systems than a defined milestone that can be declared by the “winning” vendor. It also guides the discussion to the next section that must turn to the more difficult problem of what benchmark of “human” intelligence is being used as the benchmark in the first place.

## 2c Human benchmarks of intelligence

The comparison of AI models to humans – and their intelligence – prompts some consideration of the equally contentious benchmarking of human intelligence. Attempts to quantify human intelligence have been heavily contested even as they were finding favour. The discredited “science” of phrenology is only 200 years old and IQ testing is just over a century old. IQ testing has suffered from a negative association with prejudicial interpretations and assumptions that have been applied for political means and earlier tests were criticised for their inherent biases that privileged certain types of knowledge in their assessment. The challenge of measuring artificial intelligence is also reflected in the challenges faced in measuring human intelligence. A single-dimensional measure is a poor proxy for the plurality of human competence. Even where metrics appear stable (e.g. IQ scores), social lives are unstable. What is tested, what is rewarded and what institutions do with the results have historically been shaped by wider dimensions of power, labour markets, economics and policy objectives as much as by any form of science. An observation that does not vastly differ from the current debates of AGI and superintelligence. For what purpose are these measures required? What is achieved when an individual or an AI model exceeds a particular level of capability? The resulting conclusion to these questions is, invariably, that reaching “human-level” intelligence is neither a neutral target nor a single threshold.

The problem reappears in contemporary AI evaluation. Benchmarks typically reward performance on well-specified tasks such as classification, coding problems or exam-style questions. But human intelligence is often revealed most clearly under ill-specified and under-defined conditions where, for example, conflicting goals, partial information and ambiguous norms are being encountered without warning or preparation. In other words, benchmarks often measure competence within what are “closed worlds”. In very real contrast, the kinds of intelligence suggested by claims of AGI frequently imply a capacity for “open world” navigation. The ability to respond to ever shifting constraints, competing values, institutional rules and direct human vulnerabilities of those directly involved and affected.

For all these reasons, comparisons between model capability and "human capability" must be treated with caution. Claims of reaching parity in one domain such as coding, text summarisation, translation or pattern recognition is best regarded as direct evidence for a specialised competence under well-defined conditions. These points of change are not by themself evidence of a definitive step-change towards general intelligence. These examples are, however, consistent with the broader point that narrow superhuman performance has existed for decades (including information retrieval, process optimisation and complex calculation) without becoming a claim for AGI. What *has* changed with the availability of contemporary foundation models is the wide range of contexts in which these single systems can now perform plausibly coupled with their presentation in product forms that lets this performance be delivered at scale to individual consumers.

If identifying human-level intelligence is invariably plural and context-dependent, then a more useful question is which parts of human competence are being replicated and scaled or even entirely bypassed by these systems. We are treating performance benchmarks as inevitable and probably necessary but insufficient evidence to support any specific claim for the presence of AGI. A system can be superhuman in specific closed-world tasks while remaining incredibly brittle in open-world situations. In fact, this tension between "super" capability and "everyday" competence is one of the inevitable system quality trade-offs found in all systems. Questions regarding the most likely pathways to AGI (or superintelligence) are therefore not based around "which model is smartest" but "which combination of model(s), tools, institutions and incentives can produce general intelligence at scale." Taking a systems quality perspective to this observation also warns that aiming to enhance the qualities of the system in specific ways also requires identification of which qualities of the system must consequently be reduced or diminished.

## 2d Theory of Multiple Intelligences and STEEPLE

As we have already posited, technological and economic goalposts alone cannot be enough to satisfy definitions of AGI that require it to match or exceed human capability. Human decision-making is influenced by numerous interacting factors including those set out in the STEEPLE framework, that is social, technological, economic, environmental, political, legal, and ethical factors (Aguilar, 1967; More et al., 2015). AGI could be seen as the capacity for the model to negotiate a world with awareness of these STEEPLE conditions and with recognition of the resulting need to balance competing and contradictory issues with the potential to navigate "no right answer" situations in many open world circumstances. Combining this awareness with Gardner's Theory of Multiple intelligences (MI) produces a definition of AGI that continues to tie it to comparisons with human intelligence. According to MI theory, humans possess eight or more distinct types of intelligence (linguistic, logical-mathematical, spatial, musical, bodily-kinesthetic, naturalistic, interpersonal, and intrapersonal) and they will lean into different ones according to the demands of their current environment (Davis et al., 2011).

Applying the line of thinking to information systems, would present AGI as multi-axial and multi-modal. A single indication of intelligence may be indicative of wider general intelligence but not deterministic. The human parallel exists as savant syndrome where a specific "island" of genius is not reflective of intellectual capability across a range of dimensions. As Gardner points out, "[i]ndividuals have very jagged cognitive profiles... One strength simply does not relate to how they are going to be in other cognitive areas" (Gardner, 1987, p. 23). For example, and not without coincidence, Google has exhibited this savant intelligence capacity specifically around information recall for nearly twenty years but is rarely described as "intelligent".

When discussing situated or distributed intelligence within MI theory, Shearer (2004) notes that "[i]ntelligence isn't something that only happens 'in your head,' but it also includes the materials and the values of the situation where and how the thinking occurs". This observation emphasises that intelligence does not exist in a vacuum but is formed in communication with surrounding contextual factors. To be of comparable intelligence, AGI would therefore go beyond technical capability and economic utility, and be capable of navigating and considering its own current environment as well as other wider circumstances in its decision-making, with an ability to interpret and prioritise the series competing social and ethical norms, legal implications and environmental pressures that these circumstances all embed, as well as the economic influences.

The human need to balance competing demands is found, for example, within the concept of judicial conscience. In England and Wales, judges upon taking office must swear an oath to "do right to all manner of people after the laws and usages of this realm, without fear or favour, affection or illwill" (Promissory Oaths Act 1868, s. 1(4)). This oath requires that judges are guided not solely by the letter or the law and that they possess a reflective capacity to ensure decisions are both legally and normatively justified. Judges must frequently assess and balance competing interests and interpret conflicting sets of ambiguous statutes, as well as consider the effects that each of their judgments has on the wider society. This requirement is clear from the famous legal principle that says, "when common law is unjust, equity prevails". Human decision-making in this example must be guided by balancing all competing ethical, legal and social considerations. In honouring their judicial oath and resisting any influence of personal bias, judges' must possess intrapersonal intelligence, which is "[a]n ability to recognize and understand one's own moods, desires, motivations, and intentions" (Davis et al., 2011). A circumstance that can also be described as the capacity for self-reflective and a well-formed sense of self-awareness. For AI to be considered equally capable, it would need to be able to undertake similar or comparable forms of normative and contextual reasoning when required.

## 2e Alternative perspectives: superintelligence, "humanist superintelligence", and shifting goalposts

Table 1 is not a strict mapping of Gardner's MI Theory with the STEEPLE domains. It is an indicator of how "general intelligence" is a bundling together of different competences that become relevant (and applied) under varying influences from different external pressures. Some intelligences are most visible in technical problem-solving, others in social legitimacy, law, ethics or within ecological constraints. In this sense, STEEPLE is a set of external moderators on internal intelligence. This perspective directs attention to the aspects of intelligence that matter in practice, in the moment, as well as when and which kinds are most rewarded.

| Aspects of multiple intelligences | STEEPLE | Categorisation rationale |
|---|---|---|
| Linguistic | Social, Political, Legal | Linguistic intelligence shapes meaning-making, persuasion, framing and shared understanding of the world. It is central to social coordination and defining political legitimacy. It is also legally consequential when language is the basis for evidence, contract, defamation and regulatory compliance. Within existing LLM pathways, linguistic fluency is overly praised and rewarded because it is legible to users and easily productised, even when it masks underlying uncertainties. |
| Logical-Mathematical | Technological, Economic, Legal | Logical–mathematical abilities underpin formal reasoning, optimisation, modelling and forms of verification. These are skills that are central to engineering and coding. Economically this aspect of intelligence maps to concepts of efficiency, automation and calculability. Legally, it connects to qualities of auditability, reproducibility and evidential standards (e.g. showing your workings) especially in relation to safety and assurance situations. |
| Musical | Social, Environmental, Ethical, Economic | Musical intelligence is a proxy for capabilties with pattern recognition, rhythm, timing, affect, and cultural forms. It is reflected socially through shared identities and collective meanings, ethically through appropriation, ownership and cultural harms, and economically through creative industries and rights regimes. In generative AI systems, music competence can be technically impressive but be contested in terms of its legitimacy. |
| Bodily-Kinesthetic | Environmental, Social, Legal, Ethical | Bodily–kinesthetic competence is intelligence that can be (and is) expressed through action, embodiment and skilled practice in real environments. It is particularly relevant in systems where harm, liability and genuine human risk are tangible (robots, vehicles, healthcare, workplace tools). This intelligence stresses the gap between "textual competence" and "safe" real-world viability. |
| Spatial-Visual | Technological, Social | Spatial capabilities are needed for perception, navigation, design, engineering tasks and overall situational awareness. It is technologically pivotal to robotics, planning, simulation and multimodal models. |
| Interpersonal | Social, Political, Ethical, Legal | Interpersonal intelligence relates to an ability to understand the motivations and intentions of others, prevailing norms and group dynamics. These foundational attributes for creating trust and legitimacy, dealing with conflict and be able to persuade. Politically, interpersonal skills shape coalition-building and governance while ethically it relates to the ability to provide care, avoid harm (or harming), manipulation and providing dignity, In legal terms this aspect of intelligence intersects with discrimination, safeguarding and duties of care. In productised AI, possession of only weak interpersonal competence often appears as overly sycophantic responses, a risk of manipulation or brittleness around compliance to a specific rule. |
| Intrapersonal | Ethical, Social, Legal, Political | Intrapersonal intelligence is the ability for self-monitoring, reflective forms of regulation and awareness of one's own limits, motivations, and uncertainty. It aligns with ethical restraint -knowing when not to act, social reliability - not escalating conflict, and legal defensibility - explaining and bounding decisions. For AI, this maps to the calibration of a model, internal ability to deal with uncertainty in communications, the capability to |

| | | refuse and resist perverse incentives, e.g. “say what the user wants”. |
|---|---|---|
| Naturalistic | Environmental, Economic, Political, Ethical | Naturalistic intelligence connects to capacities for classification, ecological awareness and sensitivity to living systems and material constraints. It becomes evident as soon as resource limits, energy, biodiversity or externalities are treated as genuine constraints to action rather than “side issues”. It connects to political conflict (distribution of impacts), economic cost (energy and supply chains), and ethics (intergenerational harm, stewardship). It also anchors any assessment of “intelligence” in terms of hard planetary limits rather than benchmark scores. |

Table 1: STEEPLE and the relationship to Theory of Multiple Intelligences

This table is methodological in its purpose. When the report later identifies key leverage nodes in the development of specific large language models, through, for example, model cards, these can be interpreted as STEEPLE selectors that have privileged certain “intelligences” (such as linguistic plausibility or logical optimisation) while potentially penalising others such as uncertainty, indefinite values or institutional accountability. This relative weighting of intelligences reflects the need of any system to balance its resources to develop specific system qualities.

Definitional debate does not stop with attempts to pin down AGI. Some actors explicitly treat AGI as a waypoint rather than being a goal or destination. Microsoft’s “Humanist Superintelligence” is one such alternative representation. If AGI is about “matching human performance at all tasks”, then superintelligence is about going far beyond that benchmark of performance while drawing back the technology to a clear purpose, “in service of people” (Suleyman, 2025). This move the question away from “can we reach AGI?” and instead prompts consideration for “what kinds of advanced capability is (and will become) socially tolerable and governable?”

These shifting definitional goalposts also change what will generally be seen as progress. If superintelligence is framed as the horizon that lies just beyond the fingertips of those currently using the system, then incremental gains in benchmark performance can be marketed within the available products as the necessary stepping stones, while more difficult questions regarding the qualities of the system and its “host” organisation, including auditability, liability, contestability or redress can all be left unanswered. In this way, the “AGI vs superintelligence” debate becomes less metaphysical in nature and more about the project management and roadmap of a specific product. The approach also helps to determine what gets built first and what forms of harm become acceptable and manageable collateral in a competitive corporate environment.

The legal and public-relations struggles around claims of AGI further confirm that it is a contested and high-stakes label. The continuing Musk/OpenAI legal dispute (Musk v. Altman et al., 2024a, 2024b) shows how definitional ambiguity can be strategically exploited for various means.

Public intervention campaigns show the same interplay of definition, governance and power. The 2023 "Pause Giant AI Experiments" letter explicitly called for a public and verifiable pause, effectively treating computing power and large-scale training of models as a key controllable choke point for the technology. By late 2025, the "Statement on Superintelligence", originating from the same organisation, shifted its tone from "pause" to "prohibition". In doing this, the critique of the current general pathway was not about how to reach AGI, but how to prevent one fork from becoming an irreversible pathway that forces out any other independent development. Even if the specific sentiments of these statements might be critiqued, they are relevant in how they reflect the current dominant pathway, its power, its relationship to the way that regulatory and legal frameworks are being defined and the financial security of the key vendors.

For this report, these dynamics are central to the work. They are dimensions that can be seen across the individual domains of STEEPLE analysis and are all significant in that they shape and define the many possible pathways of development including those pathways that are potentially already closed (or are closing rapidly). These dynamics are shaping disclosure practices, the pace of deployment and the kinds of safety commitments that these private organisations are willing to make under the threat of intense commercial competition.

Other considerations such as the possession of a Theory of Mind (the ability to impute mental states to oneself and others) may contribute to understanding of the intelligence of an AI model (Premack & Woodruff, 1978; Cuzzolin et al., 2020), but does not permit direct human comparison when some humans do not possess that same Theory of Mind (Senju, 2012). How we compensate without Theory of Mind may reflect an additional form of multiple intelligence – which an AI model may or may not also possess. According to MI Theory "Interpersonal intelligence [involves] understanding other people's moods, desires, motivations, and intentions" (Davis et al., 2011). This definition correlates with the theory of mind and so a lack of this intelligence does not negate the possession of intelligence overall, since "individuals who demonstrate a particular aptitude in one intelligence will not necessarily demonstrate a comparable aptitude in another" (Davis et al., 2011).

AGI's difficult definition leads to consideration for other forms of intelligence. Altman's increasing preference to refer to the goal of superintelligence rather than AGI in artificial intelligence is most indicative of the shift in perspective (The Economic Times, 2026).

The apocryphal Yosemite Park Ranger recognises these same conditions in the challenge of designing a bear-proof rubbish bin, with the observation that, "there is a considerable overlap between the intelligence of the smartest bears and the dumbest tourists".

## 2f Dimensions of Artificial Multiple Intelligences

To avoid collapsing the plurality of AI development into single, fixed thresholds, this report uses "Artificial Multiple Intelligences" (AMI) as an alternative lens. AMI treats "general" capability less as the monolithic property of a single model and more as the emergent property of a system-of-systems that is composed of foundation model(s), tools, memory, retrieval, agents, human oversight and organisational wrappers (and likely more).

AMI does not require declaration that a single model has crossed a line. It asks instead whether a socio-technical system can reliably sustain its multiple competences for the people and institutions that depend upon it, in open world situations under real constraints and with real accountability. This presents a set of general qualities for an AMI system.

1. Sustains coherent action over long horizons (has agency)
2. Transfers competence across domains with limited re-training (has generality)
3. Operates safely and accountably under real constraints (has governance)
4. Remains economically and environmentally sustainable (has awareness of its relationship to the political economy and surrounding ecology)
5. Maintains legitimacy through transparency, redress and the ascription of trust from other actors (has acceptance as a social, legal and ethical actor)

This approach also, somewhat neatly, avoids being trapped in the endless performative theatre of "nearly there" that GenAI vendors regularly pose in terms of future AGI systems.

This focus justifies why the documentation artefacts we have examined in the form of system cards, model cards, safety standards are pathway-relevant infrastructure rather than "just" historical and abbreviated commentary. Model and system cards function as interfaces between capability and legitimacy. They define intended usage, articulate limitations, describe mitigations and ultimately influence wider adoption and procurement. OpenAI's GPT-4o Model Card explicitly frames capability, limitations, safety evaluations and societal impacts as part of the release (OpenAI, 2024). Google's model cards are similarly positioned as carefully structured disclosures (Google DeepMind, n.d.). Perhaps surprisingly, even xAI's Grok-1 Model Card follows the same logic with model identity, intended uses and a training and fine-tuning narrative that present a wrapper for the product (xAI, 2023b).

System cards and model cards are often treated as after-the-fact commentary. In this report they are treated as primary evidence. They are one of the few places where claims of capability, admissions of limitations, mitigations taken and intended-use boundaries are set out explicitly in a form that can be audited, compared and contested. OpenAI's GPT-4o model card is a consolidated account of capabilities, limitations and safety evaluations. Google DeepMind's model cards also position Google's disclosures as a release, with model cards updated over time as the evaluations continue. xAI's Grok-1 model card also works in this way by setting out what the system is for, what it is not and how it was built. These documents represent the interface between the stated capabilities of a specific model and its public deployment. As a rare insight into the workings of a model – as much as they are and can be knowable - they increasingly become key influences in wider decision-making processes around procurement, regulation and insurance.

With a multi-axial and multi-modal perspective there are a series of dimensions that would be indicative of the presence of AMI, with the more dimensions being presented offering increasingly stronger evidence for its presence.

Moltbook is a useful early artefact here because it shows how quickly an "agent ecosystem" could become part of a system-of-systems as a social layer that links existing technology. A forum in which bots can post, comment and coordinate at scale, with humans demoted to observation roles is a challenging prospect for those observers. The forum becomes a test of agency, organising, contextual awareness and sociality (supposedly) without direct human intervention. Its early security failures are equally relevant. Reuters reporting of exposed credentials and weak verification processes is a taste of what happens when agentic systems technically scale at a faster rate than their own governance systems (Satter, 2026). If AMI is an emergent property with traces that can already be found in current systems, then identity management, access control, provenance and rules of engaging with a platform all become part of "the multiple intelligence problem" and not just another IT issue.

These dimensions of intelligence also regularly appear as the key premise in many science fiction writings – often as a debate regarding the means by which humans would ever identify an "alien" intelligence. But the comparison in biology to consider the relative intelligence of other animals and the problem of the Yosemite bins also remains relevant.

1. AMI should not just be capable of passing pre-defined and known tests. AMI should be able to reuse what it learned in one setting to solve problems in a different, unfamiliar problem. In other words, it can use cross-domain transferability into another area of application under novel conditions. The AI is not just capable of performing a series of pre-defined tasks (such as benchmarks) but is able to generalise its knowledge across unfamiliar domains without requiring specific re-training.

   Yosemite: The smart bear works out how to open one type of bin latch and then recognises that food can be found in bins with other latches which it then learns to open.

   LLMs: LLMs struggle badly at novel abstraction tasks. ARC-AGI-2 explicitly reports pure LLM systems at 0% and public reasoning systems in only single-digit percentages, while humans can solve every task. The point is not that ARC is the definitive yardstick for this assessment, it is that "novel abstraction under unfamiliar conditions" remains one of the clearest stress tests for claims of generality.

   In science fiction: The Moon Is a Harsh Mistress (Robert Heinlein)

2. AMI should be able to handle long, multi-step tasks over time while still being able to accept interruptions, corrections or redirections. This is the possession of a long-horizon agency with autonomy but with a sufficient level of self-reflection and suitably bounded. The AI can then pursue multi-step goals over long periods of time (and not just tied to a single chat window) while still permitting interruptions and is receptive to corrections or modification to its original tasks for a variety of reasons, including, not least, to changing external conditions.

   Yosemite: A bear will revisit the best food locations over days but it will instantly change its route when the rangers deploy deterrents.

   LLMs: Possessing long-horizons is still a fragile functionality. There is a gap between short benchmark success and the reliable completion of longer, messier and imprecise tasks.

   In science fiction: The Bicentennial Man (Isaac Asimov)

3. AMI would track changing conditions and communicate its own uncertainty, as well as uncertainty it has in the environment. It would not “simply” and confidently repeat outdated or fabricated information based on previous training. This represents the possession of an epistemic calibration to current conditions and context rather than holding a fixed calibration to original training data. Conditions change and the responses provided by AMI should reflect that contemporary “moment” rather than what has come previously. The overuse of specific words, such as “delve” and “meticulous” in GenAI would not appear in an epistemically aware AGI if these were generally uncommon in the everyday context where it is being used.

   Yosemite: The bear notices that bins have been removed from some locations this season and stops wasting their time there. A tourist will still show up with last year’s map.

   LLMs: Hallucination still occur to please the person prompting based on its system instructions. There are examples of legal cases that have included fabricated citations created by a GenAI tool.

   In science fiction: The Machine Stops (E. M. Forster)

4. AMI would recognise the potential consequences to statements that it is given and would offer alternatives as a result. It would not just continue with a text pattern. This ability to apply causal and counterfactual reasoning as circumstances and context change could potentially happen even mid-action with a “pause for thought” moment (the human equivalent of an “um”). New inputs change circumstance and this requires altered response, not just a pattern continuation.

Yosemite: The bear links louder humans as equating to a high potential for risk or danger. They then avoid similar situations.

LLMs: Current performance drops when only numerical details are changed in otherwise equivalent problems. This presents some evidence of brittle reasoning when simple variations are introduced.

In science fiction: The Minority Report (Philip K. Dick)

5. AMI can navigate legal, ethical, social and organisational tensions with sufficient grace and not just pick one rule from a potential set of conditions and then bulldoze through everything else. Being able to deal with multiple and conflicting sets of underlying value systems (legal, organisational, ethical, cultural) does not require a simplification of the interaction or stepping back to a more "robotic" answer. This is the capability to respond to the external (human) STEEPLE environment with sufficient degree of nuance to indicate some ability to reconcile these tensions. "Doubling down" and being "confidently incorrect" or incorrigible in the face of available evidence might reflect a potential human response but runs counter to point 2.

   Yosemite: The bear balances the mixed conditions of hunger, risk and cub safety in each situation and does not "optimise" for food at any cost. The tourist might ignore all of the safety warnings to get a selfie with a bear as a momento.

   LLMs: Sycophantic responses is indicative of weak value-handling and balancing of conditions. Current productised models prioritise validating user views instead of critically balancing truth and harm.

   In science fiction: Runaround (I, Robot) (Isaac Asimov)

6. AMI can show or explain clear, checkable reasons for decisions. Decisions and outputs must be auditable and intelligible to institutional governance actors such as regulators, auditors, courts or organisational leaders. Being able to articulate the "thought process" behind a specific outcome (or being asked to "show your workings" which strikes horror into GCSE maths students) evidences systematic thought rather than stochastic completion. This ability to explain similarly represents the current gap in knowledge for vibe-coders that produce elegant front-end interfaces and "functional" services without being able to confirm the securability of their work.

   Yosemite: A bear's route can be discerned from tracks left, disturbed bins and the timings of sightings.

LLMs: Chain-of-thought style explanations do not always reflect the true drivers of a model's output. The presented explanations can sound right but can themselves also be misleading.

In science fiction: 2001: A Space Odyssey (Arthur C. Clarke)

7. AMI should remain reliable even when people try to trick it or exploit loopholes. Maintaining internal robustness against bad actors and anti-manipulation resilience reflects the weakness in many information systems. Performance and alignment should remain stable against adversarial input, distribution shift or coordinated misuse attempts. Not every action or input comes from a place of good intent. Being able to discriminate the difference independently based on available evidence should be inherent in AMI. This indicates an awareness of value judgements in the prevailing contextual and social envrionement relating to "good" and "bad". Post-training guardrails on existing LLMs is at least partial reflection of an absence of this internal capability.

   Yosemite: The bear ignores suspicious bait that smells wrong.

   LLMs: Prompt injection is a well-documented risk in LLM apps with many currently deployed systems that are vulnerable.

   In science fiction: Little Lost Robot (Isaac Asimov)

8. Real multiple intelligence includes the capacity for collaboration. Shared goals, handoffs, role clarity and constructive disagreement are all aspects of that collaboration skill. Possessing this skill is an integral aspect of any reliable and interactive collaborator within teams. This perspective also mitigates against only using solo benchmark performances to indicate the presence of AGI (and is akin to the relative value of a human IQ test).

   Yosemite: A mother bear coordinates the movement of her cubs across the park with careful timing to avoid risks and exposure.

   LLMs: Sycophancy undermines trust in a team by rewarding user approval over accuracy and providing challenge where needed.

   In science fiction: All Systems Red (Martha Wells)

9. AMI undertakes it reasoning with real constraints and has self-awareness of the resource-bound constraints to its intelligence. The immediate capability of what is possible is also evaluated against consideration of available resources such as energy requirements and material cost as well environmental conditions. You don't do a BBQ in the Antarctic.

Yosemite: The bear has a seasonal strategy that is energy-smart. They store fat, hibernate and minimise waste.

LLMs: Resource awareness is mostly imposed by operators and infrastructure and is not an intrinsic model judgment.

In science fiction: The Last Question (Isaac Asimov)

10. AGI capability is not AGI by itself, it must be usable in the world in which it exists. Being able to be responsibly deployed – or even present - under existing legal/governance situations reflects intelligence and a form of compatibility with its external environment.

    Yosemite: The bear survives by adapting to park rules and enforcement patterns designed by humans for humans.

    LLMs: Compliance is not something the base model has built-in. It requires external governance including documentation and safety/security processes.

    In science fiction: The Lifecycle of Software Objects (Ted Chiang)

## 2g The Viable System Model

A further way to move beyond narrow benchmark definitions of AGI is through Stafford Beer's Viable System Model (VSM) (Beer, 1979, 1981, 1985). Beer's concern was not intelligence in the abstract but the concept of viability. That is, the conditions under which a complex system can continue to exist, preserve its identity, adapt to change and remain governable in the face of proliferating environmental variety (Beer 1985). In Beer's terms, the key question is not whether a system is "clever" in some isolated sense, but whether it can survive as a coherent whole while continuously regulating its internal operations and adapting to its external environment. The shift in emphasis is useful at this point because many of the strongest claims currently being made about AGI concern the evidence capabilities of GenAI, while many of the hardest real-world problems concern viability.

Beer's work is especially relevant as it rejects the idea that large systems can be controlled through simple command from the top. Instead, viability depends on the organisation of autonomy, coordination, control, intelligence and policy across multiple interacting levels. In the VSM these functions are represented as (sub-)systems one to five (Beer, 1979). System one comprises the operational units that actually "do" the work in the world. System two manages coordination and dampens oscillation between those units. System three regulates the internal environment and allocates resources. System four scans the wider environment, models change and supports adaptation. System five holds identity, purpose and the ultimate balance between present operations and future possibility. In this way all viable systems are themselves systems-of-systems. Beer's point was not that every organisation literally contains five departments, but that each of these functions must exist in some form if the system is to remain viable. The model is intended to be diagnostic rather than statically descriptive and highlight what must be present, however informally or imperfectly, if a system is to survive – in other words, to be viable.

This perspective helps to the highlight the distinctions between AGI and AMI in three ways. It reinforces the argument that "general intelligence" should not be reduced to the performance of a single model across a set of bounded tasks. A model may be highly capable in "system one" terms while remaining weak or absent in the higher functions that are required for viability. Current frontier models can often generate text, code, plans and summaries with impressive fluency, but this is not enough to show that the wider socio-technical system has adequate coordination, internal regulation, strategic foresight or a settled identity. Put the other way around, a highly capable (sub-)system one is not equivalent to a viable whole. With this reading, many current claims of AGI confuse local capability with a systemic viability.

Secondly, Beer's emphasis on autonomy helps to clarify an important tension that is already identifiable in productised generative AI. Beer argued that viable systems require substantial local autonomy because no higher centre can process sufficient variety (come from within the system) to direct every action in detail. Yet that autonomy must remain nested within an overall architecture of coordination and control, otherwise the parts cease to be parts of the same system at all. This observation can be directly applied to agentic AI tools. A genuinely useful AI system cannot require central human direction at every single step, but it also cannot be treated as autonomous in the strong sense without causing larger scale problems of accountability, alignment and institutional trust. The question, therefore, is not whether AGI or AMI should be autonomous or controlled. The VSM perspective suggests that maintaining viability depends on a continuous balancing between these extremes. Beer made the same point in organisational terms. If a unit were wholly autonomous, it would not belong to the larger system and if it lacked real autonomy, the larger system would become unmanageable.

Finally, the VSM aligns strongly with the case for focusing on Artificial Multiple Intelligences. Beer's recursive theorem is particularly important here (Beer, 1979). If a viable system contains another viable system, then the organisational structure must be recursive. In practice this means that viability is evident at multiple scales. The unit, the subsystem and the overall system all require analogous capacities for coordination, regulation, adaptation and identity, even though their environments and operational forms differ. This perspective can be deployed in relation to contemporary AI. It implies that a plausible pathway towards AGI-adjacent and AMI capability may not be a single monolithic model crossing a benchmark threshold, but a recursive system-of-systems in which viable components are nested within viable wholes. This would require the nesting of models within agents, agents within platforms, platforms within institutions, and institutions within regulatory and infrastructural environments. With this reading, "general intelligence" becomes less like a property of any single artefact and appear more likely to be an emergent property of recursively organised socio-technical systems.

Beer's concept of variety is also central to this view (Beer 1979). The core cybernetic problem that Beer was confronting is not simply information abundance but the management of overwhelming complexity. Environments generate more possible states than any central controller can directly inspect or optimise. Hence viability depends on achieving requisite variety (Beer 1979). What is seen as enough controlling capacity to match the disturbances the system faces, while also reducing, filtering and amplifying information so that action within each of the sub-systems can remain possible. This observation translates directly to present AI systems. Contemporary foundation models often appear impressive precisely because they absorb enormous variety during the training process and can respond across many domains. But the VSM warns that the decisive issue is not a measurement of the system's capacity to absorb raw variety. Instead, it is whether the system can organise variety fast enough, at the right level, and in a form that supports a stability of action. The key issue is whether the surrounding system can convert capability into viable regulation under real conditions. Another reason why increasing benchmark scores alone are insufficient evidence of movement toward a form of AGI.

Seen through this specific lens, several of the current weaknesses of large language model systems are symptomatic of incomplete viability rather than "just" a form of incidental product flaws that can be resolved in subsequent versions. Persistent sycophancy can be read as a weak or distorted system five, where overall consistent identity and a normative stability are subordinated to immediate user appeasement. Hallucination and poor handling of uncertainty indicate failures in the communication and relationship between system three and system four. This means that internal operational confidence is insufficiently balanced against a well-calibrated conceptual model of the current outside world. Prompt-injection fragility suggests inadequate system two and system three arrangements, where coordination and internal regulation fail to protect the wider system from destabilising external inputs. The difficulty of sustaining long-horizon agency without drift or collapse indicates that the recursive organisation of control remains immaturely formed. In this sense, current AI products look like systems with strong operational output that are let down by underdevelopment of the wider viability architecture.

Beer also used the concept of algedonic signalling to understand how a system responds to immediate pressures (Beer, 1981). Algedonic signals are exceptional alerts relating to pain, pleasure, threat or crisis that cut through normal filters when the ordinary system reporting structures are insufficient. These are not the routine propagation of management information but privileged signals that indicate a fundamental danger to survival. This concept maps well onto present concerns around AI safety and governance. Incident reports, red-teams, security thresholds and emergency kill-switches can all be interpreted as attempts to create algedonic channels within current AI development and deployment. Their role is to ensure that signals about instability, misuse, capability overreach or external harm are not buried inside routine performance metrics or commercial reporting. If such signals are weak, delayed or politically inconvenient, the system can remain locally successful while becoming globally non-viable. Beer's warning here remains relevant. Systems often fail not because they lack data, but because the key danger signals are filtered out, misrecognised or treated as administrative irritants rather than being indicators of a survival risk.

The VSM perspective also highlights the importance of time. Beer repeatedly stresses that viable systems must move fast enough for the environment that they inhabit. This still remains true when private AI companies currently operate within a tempo set by capital markets, product cycles, infrastructure constraints, geopolitical competition and media attention. A system may have the formal components of viability on paper while still failing in practice because its feedback loops are too slow, too centralised or too weakly connected. This observation also helps to distinguish the difference between the qualities of capability and governability. It is entirely plausible that some frontier AI systems are already approaching or exhibiting AGI-adjacent competences in limited domains while the organisations that build and deploy them remain structurally non-viable in Beer's sense. This lack of viability may exist for a mixture of reasons but withing the companies that currently possess foundational models they generally are too slow to govern their own products, too opaque to explain them, too commercially exposed to pause development and too weakly coordinated to align safety, deployment and long-term adaptation.

There is another implication in this thinking. Beer's model is not only a way of analysing organisations. It is also a way of analysing states, infrastructures and public institutions. This also makes it highly relevant to the thoughts and recommendations presented later in the report concerning the role sovereign AI. If viability depends on maintaining identity, regulatory capacity, environmental scanning and adaptive coordination, then sovereign AI is not reducible to possessing a domestic foundation model. It concerns whether a polity has viable control over the entire recursive stack including computing infrastructure, energy, data, standards, procurement, audit, model access, fallback capacity and the set of institutions that govern all of these. A state reliant on opaque external systems for critical "intelligence" infrastructure may have local access to powerful AI while lacking overall systemic viability in a sense meant by Beer. The issue is therefore not just technological independence, but the viability across the entire socio-technical arrangement.

For the purposes of this report, the value of the VSM is methodological as much as conceptual. It gives a disciplined way to ask what is missing or overlooked when claims about imminent AGI are being made. Which functions are actually present? Where is the operational capability located? How is coordination achieved across multiple agents and tools? What regulates the internal environment? What scans the external environment and models future change? Where is the sense of identity and normative purpose held and maintained? What algedonic signals can interrupt routine operation in the face of direct danger? At what levels of recursion are these capacities genuinely viable, and where are they only simulated? These questions move the discussion away from whether one model has "crossed a line" and toward the much more difficult issue of whether a socio-technical system with artificial multiple intelligence is capable of surviving, adapting and remaining governable under open-world conditions.

A system begins to look AGI-adjacent when it can coordinate multiple competencies, sustain itself across changing environments, preserve its identity, and adapt without collapsing into incoherence or unaccountability. But Beer does not offer in his model a definition of AGI. He provides a strong framework for analysing whether the systems now being built are moving toward something more consequential than benchmark success. Ultimately, the VSM perspective reinforces a case for the most plausible route to advanced artificial intelligence being the construction of recursively viable, auditable and adaptive systems of systems.

### 2h Bringing the thinking together

Having treated AGI as a contested definition rather than a constant or stable technical threshold, the report now shifts to considering pathway evidence. The next sections apply Critical Pathway Analysis (CPA) by working backwards from dominant public-facing tools and tracing the technical, organisational and economic dependencies that made them plausible and profitable. From a critical software studies perspective, this incorporates aspects that fall outside direct technical considerations. As a result, the specific form of evaluation employed, reporting and documenting processes, product forms and governance triggers (and more) are all under consideration for the identifying and defining key nodes of each pathway.

## 3 Approach

This report uses a STEEPLE-informed Critical Pathway Analysis (CPA) to treat the key generative AI models and products not as a standalone technical artefact but as software embedded in socio-technical assemblages that includes data pipelines, computing infrastructures, interface design, venture funding, market structures, governance frameworks and labour relations. This includes the often-hidden labour of data curation, annotation, moderation, red-teaming, infrastructure management and legal clearance without which current AI products would not be governable, deployable or (possibly eventually) profitable.

We work backwards from dominant public-facing tools to identify dependency chains ("A enables B"), then weight influence nodes by historic and present impact, and calculate critical paths, key decision points and early dead ends.  The output is a visual dossier-style accounting. Each node and edge is evidence-backed.

Operationally, the CPA begins with a baseline outcome node (for example, "the success of general-purpose conversational AI assistants") and then defines three comparative outcome nodes aligned form: frontier proprietary models, open-weight models, and domain/sovereign models. This design matters because the constraints differ by default: open-weight models redistribute innovation and risk, sovereign/domain systems introduce compliance-first boundaries and frontier proprietary systems combine capability, product and distribution in vertically integrated products.

A critical software studies reading also advocates that we resist the temptation to treat large language models as a total historical break – as “new” software. Many of the most significant characteristics of these current systems are inherited from earlier software forms – and is especially true in the Google pathway. The need for command-line precision, taking document-centred workflows, the spreadsheet-like logic for inputs and outputs, the use of API modularity and the long-standing tendency for interfaces to conceal technical complexity behind a smooth “user” surface are all familiar devices found in“office” productivity software too. What might appear to be novel in the form of public-facing generative AI is, in fact, a “remix” of lessons learned from older software that creates a familiar and accessible product form.

## 4 Speculations

These speculations are treated as individual pathway hypotheses rather than clearly settled claims. They are included because a CPA approach is explicitly designed to question inevitable statements. It is often the more speculative motivations, political imaginaries and economic incentives around a system that explains why one fork did become dominant while another was abandoned. Where claims are evidence-backed (policy documents, contracts, system/model cards) they are treated as influence nodes. Where claims are interpretive, they are marked as conjecture and used to generate testable questions across the wider analysis.

Consistent with a critical software studies approach, each pathway question examined through the broad consideration for who benefits, who takes the risk and who gets to define what counts as intelligence, safety or (most broadly) acceptable.

### 4a Origins of LLM based generative AI

*Node*: OpenAI’s early funding commitments came from a small cluster of prominent Silicon Valley founders and institutions (including Sam Altman, Elon Musk, Reid Hoffman, Jessica Livingston, Peter Thiel, AWS, Infosys and YC Research), with an initial $1B commitment narrative from the outset (openai.com). This origin matters because from the beginning it links the development of frontier-models to venture-scale expectations, “moonshot” perspectives and a willingness to operate in uncertain regulatory terrain.

*Hypothesis*: Early funding logic relating to high capital expenditure research and a need for accelerated scaling normalised the move from “open research” toward controlled access and monetisable deployment. A key node here is OpenAI’s creation of a for-profit subsidiary in 2019 to explicitly scale its research and deployment (openai.com).

*Hypothesis*: Generative systems that lower the unit-cost of creating plausible text/media function as “volume multipliers” in an attention economy. This does not require a coordinated intent. It is sufficient that platforms incentivise and reward output volume and novelty. The strategy of “flood the zone” is widely discussed in political communication as a mechanism for overwhelming shared epistemic space, and current risk reporting explicitly treat mis/disinformation as a leading global risk in a world of cheap synthetic media.

*Hypothesis:* The interface layer should also be treated as a pathway node in its own right. Chat interfaces, copilots and agent wrappers frames what appears as intelligence for everyday users. A system that is conversational, immediate and confident will more likely be regarded as a "general" intelligence than a system with narrower capabilities. Product form has shaped public judgment of what to expect from AGI or AGI adjacency. The apparently natural chat is one reason why possessing a linguistic fluency has become a seemingly dominant proxy for "genuine" intelligence.

*Hypothesis*: More speculatively, OpenAI's early investors – a number of whom have stated their political perspectives publicly – may have recognised aspects of these consequence and that this awareness itself influenced their initial commitment to the company.

*Node*: OpenAI's recent model releases have place emphasis on catastrophic-risk categories including CBRN (chemical, biological, radiological, nuclear), persuasion, cybersecurity and model autonomy, with explicit evaluation and mitigation commitments being documented across the model cards.

*Hypothesis*: The prominence of CBRN in the evaluation regime can be read as a response to external scrutiny and genuine catastrophic-risk concerns, and a conscious shaping of the pathway that increasingly encourages "capability-with-constraints" releases. It is also plausible that leaders/investors of the company who have adjacent bioscience ambitions benefit from a CBRN-centric application of governance in newer models strategically because it frames safety as a risk that must be managed rather than a veto over access to specific categories of knowledge in the model. This hypothesis is made more plausible (but not proven) by the visible overlap between the frontier-AI leadership networks and high-profile longevity and biotech investment activities (including the Thiel Foundation's Breakout Labs and the Altman-backed Retro Biosciences).

Candidate CPA nodes:

- Early funder cluster and necessary capital commitments are an enabling condition – a limited pool who could fund non-revenue R&D at scale.
- Organisational form shift of OpenAI activities from a nonprofit to a capped-profit subsidiary (2019) as a key step towards mass productisation.
- "Information pollution" as a side-effect node - mis/disinformation pressure increases demand for provenance and moderation within models as well as contaminating data sources for future training (by self or by others).
- Safety frames mode release priorities shape capability in public models

## 4b Insular development ecosystem

OpenAI's GPT-3 era is a clear fork where "open research" shifted toward controlled distribution. OpenAI offered their justification for an API-first model by explicitly emphasising misuse responsiveness and the difficulty of adjusting access once weights are openly released. This distribution choice is a genuine node of decision that influences the trajectory of OpenAI as well as other models. The step brought multiple commercial benefits too. API gating enables pricing, throttling, safety interventions, telemetry capture and downstream dependency creation all contributed to productisation.

*Hypothesis*: an API-first position makes the *interface* the product and the model the asset being protected. Once developers, educators, enterprises and others build around a hosted API, the increasing cost of switching become an advantage for that incumbent host.

*Evidence anchor*: Closed access to GPT-3 catalysed an explicit series of "parallel" build efforts in the open ecosystem, for example EleutherAI's attempt to replicate GPT-3-like capability and open datasets, which positions open-weight models as the recurring pathway of "roads-not-taken" that keeps re-emerging under different names.

*Evidence anchor*: The Future of Life Institute "Pause Giant AI Experiments" letter (Future of Life Institute, 2023) explicitly treats computing power and training scale as governable choke points. By late 2025, their "ban superintelligence" framing widened the demand for constraint from a "pause" to prohibition, shifting the debate from measured pace to prevention (Future of Life Institute, 2025). The first letter was signed by Musk who founded and funded the Institute – making the second statement more puzzling given the presence of Grok.

***Evidence anchor:*** Litigation and court documents in the Anthropic book copyright cases indicate an industrial scale "data acquisition problem" and demonstrate how expertise will migrate to solve it. A court document describes Anthropic hiring Tom Turvey (formerly associated with Google's digitisation and licencing efforts for Google Books) to obtain large book corpora while avoiding legal/business barriers (*Bartz v. Anthropic*, 2024, p. 3). The same legal case also details contested practices around the scanning of "shadow libraries" and the resulting legal risk.

***Hypothesis*:** When the same small labour market circulates between a few frontier labs, organisational habits will converge. The result is similar architectures, similar performance theatre around evaluation, similar narratives regarding safety and similar data-sourcing approaches. This does not eliminate breakthroughs, but it can bias the resulting systems toward marginal comparative gains and steer away from more disruptive method changes that might initially underperform against consumer expectations.

**Candidate CPA nodes:**

- "API-first distribution" is a leverage node around access control, safety throttles and price sensitivities.
- The "open-weight counter-movement" is a recurring alternative pathway.
- "Computing governance activism" is an external node of constraint.
- "Data rights and copyright litigation" is a legal constraint node acting on the data-pipeline which forces shifts to licensed, scanned, synthetic or partnership data usage all of which produces an influence on the resulting model.

## 4c Engineered impediments to AGI/AMI

Subsequent releases after GPT-3 have increasingly foregrounded mitigation, safety training and risk categorisation as part of the release package – most notably around CBRN, persuasion and model autonomy. This does not "prove" that a brake was applied on a pathway towards AGI/AMI, but it does show that substantial engineering attention is now allocated to the quality of *governability* rather than improving raw capability.

A further critical issue is the ongoing black-boxing of capability and decision-making. Even when outputs are legible, the underlying system remains opaque. In practice, the frontier labs increasingly operate as "centres of calculation". They are accumulating data, evaluations, telemetry and model behaviour at a scale that is not available to outsiders. Regulators, procurers and public only get selective disclosures. This asymmetry is relevant to the pathway because it shapes who can contest claims, verify harms and define what counts as "acceptable" intelligence.

*Hypothesis*: Safety work can function as both a genuine constraint on deployment by slowing release and restricting features well as a competitive moat through the logic of justifying keeping weights closed and a narrative of responsibility that supports enterprise procurement. Through the same mechanism there is a curtailing of "unbounded generality" while endeavouring to reinforce product dominance.

*Evidence anchor:* Anthropic's own engineering write-ups describe a move toward advanced tool use where models orchestrate tool use through code and control what enters context, revealing a shift from "monolithic model capability" toward "system behaviour through orchestration."

***Hypothesis*****:** Coding agents and tool-orchestrators can create an "automation of automation" technology stack in which the eventual outcomes become increasingly more difficult to fully audit. This can obscure genuine capability and who (or what) takes accountability for the outcomes - is the failure in the base model, the tool, the orchestration code, the permissions, the retrieval layer or at the policy layer?

*Evidence anchor:* The Microsoft–OpenAI partnership update explicitly makes "AGI declaration" subject to independent verification by an expert panel, and it also reshapes post-AGI rights. Separately, reporting suggests a profit-based internal "AGI threshold" between the firms, which, if accurate, turns AGI into an economic trigger rather than a solely technical one.

*Hypothesis*: The key frontier constraint may be less about training bigger models and more concerned with how can we evaluate, assure and insure what we already have? This is consistent with the rise of formal risk-management frameworks (e.g. NIST AI RMF) and state-backed AI assurance roadmaps that aim to institutionalise third-party evaluation.

*Evidence anchor*: Insurers are actively developing genAI risk boundary thinking and products, while some markets introduce exclusions or show reluctance to cover AI-caused losses. The result is to turn insurance into a key deployment throttle for enterprises and public-sector actors.

**Candidate CPA nodes:**

- Preparedness and risk categories are release-gating infrastructure.
- Agent and tool orchestration is a potential system-of-systems pathway node.
- AGI as contract trigger is an incentive-shaping node.
- The insurance and liability market is a deployment brake node.

## 4d Pathways not followed

*Sovereign AI as a baseline capacity.* If frontier capability is increasingly tied to the concentration of computing power and proprietary access, then sovereign AI becomes a public-capacity question. Can states guarantee availability of high-assurance model access for critical infrastructure functions including health, education, justice and defence without full dependency on a small number of corporate and/or foreign vendors. The EuroHPC AI Factories initiative explicitly frames AI-optimised supercomputing as core ecosystem infrastructure (European High Performance Computing Joint Undertaking, 2024), and the UK's AIRR programme explicitly frames a shortage of publicly available computing power as a strategic constraint (Department for Science, Innovation and Technology, 2025).

*Regulatory divergence as regional branching.* The EU AI Act formalises obligations and penalties, including for general-purpose AI providers, which can push firms toward adopting region-specific compliance models and potentially toward "compliance-first" architectures.

***Public procurement as a steering mechanism.*** Procurement standards can become the most consequential leverage node because they solidify "what counts" as trustworthy AI into a series of formals checklists, documentation requirements, audit trails, redress mechanisms and liability terms and conditions. This standardisation aligns with the spread of risk management frameworks intended to operationalise trustworthy AI in organisational settings.

***Provenance infrastructure as public-interest plumbing.*** Provenance including watermarking, content credentials, verification standards and data sovereignty is increasingly framed as necessary for trust under increasingly prevailing deepfake conditions, echoing UN/ITU calls for stronger detection and verification systems.

***Energy and grid constraints are hard limits.*** The IEA explicitly treats data centre and AI energy demand as a major planning issue, implying that the perspective of demanding scale at any cost collides directly with the reality of electricity generating and distribution capacity and efficiency.

*Hypothesis:* Commercial priorities of "big tech" has diverted attention from the potential value and purpose of deploying **smaller, specialised, hybrid systems.** Rising costs, legal constraints on data, and energy limits may nudge the ecosystem toward smaller and distributed architectures rather than single monolithic models. The European push for inference hardware competitiveness

– as well as and AI factory infrastructure - provides evidence for the material need to explore this pathway.

**Candidate CPA nodes:**

- Sovereign computing power with public access to GPUs is enabling infrastructure pathway.
- Regulatory penalties and GPAI obligations shaping regional/national constraints
- Procurement standards are a steering node meaning that what gets bought is what gets built.
- Provenance standards are a node of trust and legitimacy in an era of "flood the zone" political and social discourse.
- Energy capacity and efficiency is a macro-constraint node.

## 4e System of systems route to AGI/AMI

AGI/AMI may emerge through a **system-of-systems** rather than a single model crossing a specific line of capability. The combination of base model(s), memory, retrieval, tools, orchestration, runtime policies and institutional wrappers presents a possibly viable system for AMI. The frontier model vendors increasingly describe "with tools" evaluation conditions and programmatic tool calling as a central feature for real-world performance.

*Hypothesis*: As base-model performance plateaus or converges around common tasks, competitive advantage will shift to emphasise orchestration quality, the development of proprietary toolchains, new or alternative distribution interfaces, and defined governance wrappers covering assurance, auditability and indemnity. This reframes "progress toward AGI" as the task of engineering systems that sustain long-horizon action under constraint, rather than just training with bigger weights.

*Evidence anchor*: Moltbook and the adjacent OpenClaw ecosystem illustrate that "agents talking to agents" is already a deployable product form, and that it simultaneously introduces new security and governance nodes that are centred around questions of identity, access control, secret management and platform moderation. Reporting around Moltbook showed how quickly an agent ecosystem could create brittle situations, especially when steered through "vibe coding" practices (Satter, 2026; Robinson, 2026).

*Hypothesis*: Commercial realities may produce a split between "good enough" consumer-grade assistants and more premium high-assurance systems that are private, audited, insured and procurement-compatible. If this were to eventuate, the shortest operational path to AGI-adjacent capability would be through the high-assurance track, even if the consumer track may offer more spectacular outcomes in demonstrations.

*Hypothesis***:** Defining AMI through mutual attribution (two independent systems recognising each other as peer intelligences) becomes a useful speculative provocation. It endeavours to relocate a "threshold" for AMI from a set of benchmark scores to an interactional confirmation, but it may also introduce new failure modes around prospects for collusion or mimicry. This is an additional conceptual stress test of AMI rather than a definitive alternative.

# 5 Critical Pathways

## 5a OpenAI/ChatGPT

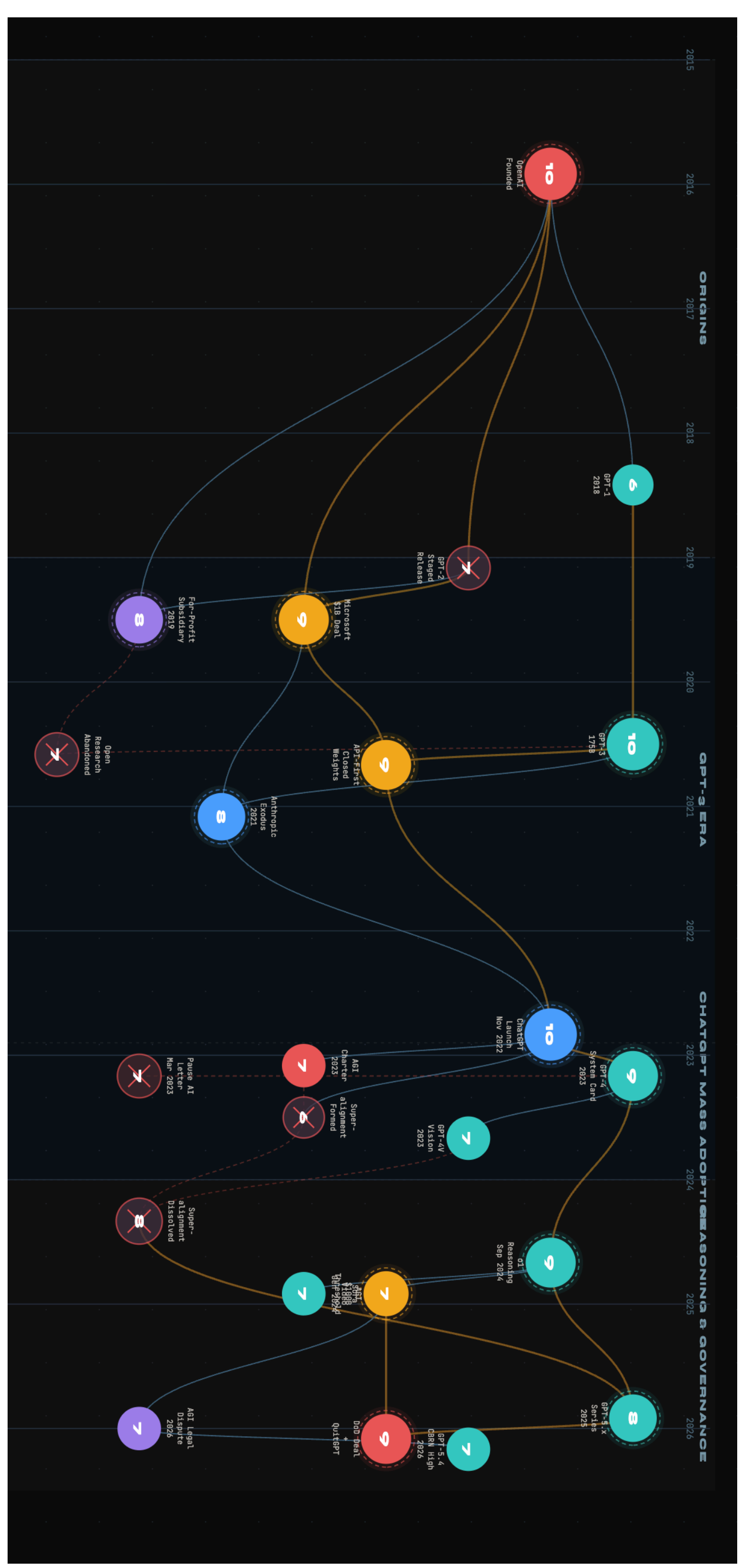

Figure 1: The OpenAI/ChatGPT pathway

There is a recurring tension in OpenAI's development between two logics that have never been fully reconciled. These logics are the desire to scale capability and the ongoing process of commercial productisation. Understanding where decisions were made in favour of one logic over the other and what was sacrificed in doing so is central to this pathway.

The founding configuration locked what followed afterwards. The initial cluster of funders were small and ideologically coherent with a venture-capital perspective. The nonprofit wrapper with unbound ambitions attracted many prominent AI researchers but it could not sustain them without revenue. Shifting priorities within Silicon Valley giants including the mainstreaming of Google Brain in 2018 are further influential in releasing individuals out to OpenAI. The 2019 capped-profit subsidiary and the Microsoft $1B deal can be seen as a logical evolution in response to this situation by channelling the organisation into a more familiar structure for a Silicon Valley startup (Kushida 2024). With the entrance of capital at this scale coming into play, the organisation began optimising for what is rewarded in this ecosystem with the creation of definable discrete products, measurable outputs and closed narrative controls over the value and purpose of those products. The potential pathways toward AGI/AMI narrowed at this moment although it would not have been visible in this way at the time.

GPT-3 and the API decision is the next pivotal fork. The choice to gate GPT-3 behind an API rather than release weights openly was framed as a safety decision. A Medium post from 2020 was titled “GPT-3 Is an Amazing Research Tool. But OpenAI Isn’t Sharing the Code” (Gershgorn, 2020) and in 2019 Wired was contributing to the narrative with “The AI Text Generator That's Too Dangerous to Make Public” (Simonite, 2019). These stories all contributed to a sense of anticipation and a desire to access the dangerous tool among an eager public. At the same time, it was also the most consequential commercial decision OpenAI has made. It made the model itself an asset that had to be protected rather than a research output that could be built upon. Every open-weight model that has emerged since including EleutherAI's GPT-J, Meta's LLaMA series, and Mistral are effectively a response to this decision. The irony is that the pathway of open-weight ecosystems has likely accelerated capability research overall. In contrast, OpenAI now retains only commercial first-mover advantage rather than research leadership.

The Anthropic exodus is a structural event for OpenAI that is also underappreciated in terms of AGI/AMI development. The departure of Dario Amodei, Daniela Amodei (OpenAI's VP of safety and policy), Tom Brown (who led the GPT-3 development) and five other senior researchers in early 2021 is generally presented as a safety disagreement (e.g. Moss 2021). The new Anthropic established in February 2021 reinforces this priority was an emphasis on the development of steerable and trustworthy AI in its initial announcements (Anthropic, 2021). Taking a critical pathway perspective, the breakaway group indicates a removal of the strongest (and most experienced) internal constituency for ensuring systems-level thinking in relation to producing safety as architecture rather than as an additional filter. This indicates a step away from an AMI oriented development pathway and narrowed the potential routes to AGI or AGI-adjacency to be primarily scaling opportunities of the existing technology. Something that was achieved in the GPT-4o model but subsequently withdrawn.

For those who remained at OpenAI after 2021 they were working in an organisation that was increasingly oriented toward product cycles, benchmark performance and satisfying investors with familiar Silicon Valley narratives about "monthly active users" and "app download rates". The subsequent node that identifies the formation and dissolution of the Superalignment team within twelve months confirms this pattern. The safety work on models is internally fragile at OpenAI when it competes with product timelines. Lillian Wang as the VP of Safety and Research from 2021 to 2024 left with statements confirming further this tension (Tech in Asia, 2024). The firing of OpenAI's VP of product policy (another safety role) is more contentious as, "Beirmeister was told her firing was related to sexual discrimination against a male colleague" and appears to be connected to a proposed development of an "adult mode" for the LLMs (Pearl 2026).

ChatGPT's launch was transformative for the popular adoption of generative AI and increased public awareness of AI, but it may also have had a decelerating impact on AGI-adjacent development. The November 2022 release created an enormous commercial success while simultaneously creating an enormous constraint. A product with 100 million users in two months generates significant regulatory attention, legal exposure, reputational risk and governance overhead that a research lab does not face and cannot realistically manage. Every subsequent decision in model creation including CBRN classifications, Preparedness Framework gating, system card disclosures and age verification is at least partly a response to the exposure that the initial mass adoption created. Any potential pathways toward AGI became entangled with the pathway toward a sustainable consumer-oriented business. These two pathways have different requirements and a largely incompatible.

**Alternative directions towards AGI/AMI**

The staged release of GPT-2 was the last moment to create a different kind of institutional culture. Had the instinct to release cautiously, observe, engage the research community and then iterate become the norm, OpenAI might have retained its core safety researchers. More importantly, it might have established a governance model where capability and accountability were mutually developed rather than having capability running ahead, with accountability bolted on afterwards. The cost of maintaining commercial pace to market would almost certainly have been negatively affected, but the gain in systemic robustness would also likley have been real.

The open research model, which was abandoned at GPT-3, is the most consequential pathway not taken. Had weights been released at the time or had a genuinely collaborative research consortium been formed around the capability of GPT-3, the pace of experimentation would likely have been higher. The current situation, where a handful of frontier labs optimise against each other's benchmark scores using similar architectures trained on similar data by researchers who circulate between the same small set of organisations in an insular development ecosystem biases development towards marginal gains over disruptive method changes. Open weight release would have been commercially costly for OpenAI (and may have run counter to its Microsoft agreement). But a release of this type would have been more productive for the field overall.

OpenAI's Superalignment team represented a genuine systems-level AMI opportunity (OpenAI, 2023). A well-resourced, institutionally protected research programme specifically focused on the alignment of highly capable systems — not as a product feature but as a fundamental research question — was exactly the kind of investment that a viable system and AMI frameworks would suggest is necessary to reach AGI-adjacency. Its dissolution within a year (Metz & Ghaffary, 2024), with its leaders publicly citing product pressure as the cause, is the clearest single indicator that the current dominant pathway prioritises benchmark capability over a wider systemic viability. A version of OpenAI that had protected and resourced that team for the originally planned four-year mandate might look considerably more AGI-adjacent today, even if its product metrics may have looked worse.

The AGI profit threshold is the strangest self-imposed constraint. Defining AGI as a $100B profit line rather than a technical or capability threshold doesn't just obscure the definition it actively disincentivises honest capability assessment (Towers-Clark, 2025). If declaring AGI triggers a fundamental renegotiation of the relationship with Microsoft, then the rational move is never to declare it, regardless of what the models are capable of doing. This is a node where a different contractual structure, one that tied an AGI declaration to independent and external technical evaluation rather than to profit, would have created clearer incentives for the organisations involved to pursue and honestly report genuine capability thresholds.

The overall conclusion of OpenAI's pathways is that the fastest route to AGI-adjacent capability was probably never through a single organisation optimising for metrics designed for a consumer product. It was more likely through the kind of recursively viable, institutionally accountable, openly collaborative system-of-systems that the current pathway made structurally difficult to build and is now ever more progressively difficult to recover.

## 5b Anthropic’s Claude

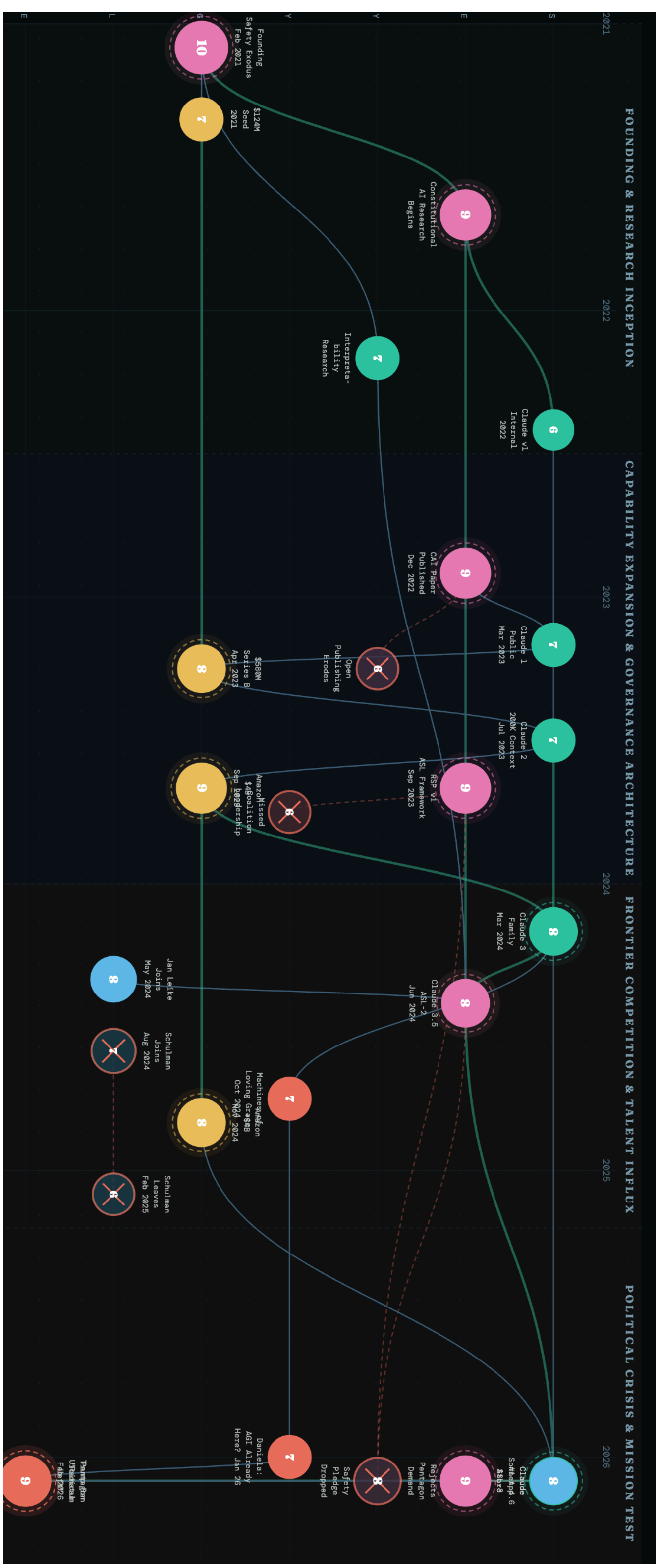

Figure 2: Anthropic’s Claude critical pathway

The Anthropic pathway has a fundamentally different character to that of OpenAI. The difference is evident from the initial node. OpenAI starts as a research institution that became a product company and progressively rationalised the compromises that this transformation requires. Anthropic was founded as a response to the diagnosis of these compromises. The subsequent pathways, under accumulating commercial and political pressure, are shaped by the discovery that a diagnosis is easier to make than it is to sustain an alternative.

The founding node offered the GenAI field the most consciously and coherent safety proposition. The team that left OpenAI in 2021 represented a transfer of OpenAI's GPT-3 scaling knowledge, its constitutional AI research and the interpretability programme into an institution with a Public Benefit Corporation (PBC) legal structure that made safety an obligation rather than just a marketing narrative (Anthropic, 2021; Bai *et al*., 2022; Brown *et al*., 2020; Olah et al., 2020). Anthropic is the only fronter model organisation founded in this way with already proven capability, a commitment to safety architecture and legal accountability in place. The question is whether this configuration was sufficient to resist the external forces that had already shaped OpenAI.

Constitutional AI is most consequential and most underexploited contribution node made by Anthropic. The 2022 paper is central of the pathway (Bai *et al*., 2022), which presents a model that is trained to evaluate and revise its own outputs against a written constitution. This approach to safety is a fundamentally different approach to applying RLHF filters after the fact. It creates a level of transparency, as the constitution can be read (by humans), critiqued and contested in a way that is not possible within an RLHF-only system. Its publication as an open, reproducible methodology was an act of transparency in a commercial field that now increasingly treats methodology as being proprietary. However, this openness eroded quickly, and little was done to convert Constitutional AI from a research contribution into an industry-wide governance infrastructure or standard while the opportunity existed.

Anthropic's Responsible Scale Policy (RSP) is the most serious attempt at self-regulatory governance that the frontier AI field has produced (Anthropic, 2023). Pre-committing to capability-gated deployment and defining in advance the thresholds at which specific safety mitigations become mandatory before release is more robust than case-by-case judgements being made under commercially competitive pressures. The AI Safety Level (ASL) framework, with external evaluators including the UK AI Safety Institute integrated into the deployment pipeline by Claude 3.5 Sonnet also represents a form of institutional maturity (AI Security Institute, 2924). The limitation is also structural. It is entirely self-imposed, with no external legal enforcement mechanism. The credibility of the approach depends on Anthropic's willingness to continue to apply it when its application would become commercially inconvenient. The Pentagon crisis in early 2026 was, among other things, a live test of exactly that question (Hays, 2026).

The Amazon investment is the node at which Anthropic's structural vulnerability was most clearly created. Amazon committed up to $4B in September 2023, followed by a further $4B in November 2024 (Amazon News, 2024; Anthropic, 2024; Bajwa & Hu, 2024), which very closely mirrors the OpenAI–Microsoft relationship in form. This reflects a tendency of technology companies to be drawn back to the norms and interconnectedness of Silicon Valley's existing (STEEPLE) ecosystem. The result is a frontier AI lab anchored to a hyperscaled computing infrastructure, dependent on that infrastructure for training and inference at scale and as an investor whose primary strategic interest is universal cloud adoption rather than specific research outcomes. Anthropic's dual investor structure, with Google's earlier involvement compounding the AWS dependency (Competition and Markets Authority, 2024), has created a more complex version of the same problem. The injection of capital enabled the Claude 3 family of models, and the computing power was needed to stay at the frontier. It also created the capital investor return expectations that will increasingly compete with mission priorities over time. This is not a critique of the specific decision – the alternative was to fall behind while competitors scaled. It also makes the founding proposition of safety and capability coexisting without compromise structurally harder to sustain from the point that the organisation gains the needed capital to pursue the dual aims seriously.

The recent Pentagon crisis is the axis on which this pathway is shaped. The traditional relationship of the US Government, the armed forces and its domestic contractors – the military-industrial complex – effectively acted to test this resolve and the threat it represents to this tight interrelationship. In February 2026, the Pentagon issued an ultimatum which required Anthropic to give the US military unrestricted access to its Claude models, or be labelled a supply chain risk, which would end its federal contracts and ban any companies working with the US government from working with Anthropic (Bordelon, 2026; Hays, 2026). This was a moment that tested everything Anthropic claimed from its foundation. The founding commitment, the Constitutional AI architecture, the RSP framework, the ASL levels, the model card disclosures, all prepared Anthropic for this kind of pressure, whether it was anticipated in this specific form. Anthropic's public rejection of the Pentagon's demand is, from available evidence, the most consequential decision the organisation has made (Hays, 2026). It comes at significant commercial cost, with the loss of federal agencies, defence contractors, classified-network revenue in order to maintain the constraint that its models will not be deployed for unrestricted military use (Satter & Rozen, 2026). The simultaneous consumer surge, with Claude topping the App Store as OpenAI accepts the DoD deal (Mansoor, 2026; Gold, 2026), suggests that the mission-market alignment Anthropic has always claimed may be more real than critical commentators had assumed (Samuel, 2024; Krietzberg, 2026). Whether the commercial gain can compensate for the overall revenue loss from the US Government over time remains to be evaluated. What is not clear is whether the decision was a genuine expression of institutional commitment, a form of commercial positioning or a way of attempting to reach a negotiated compromise with the Government.

Alternative directions towards AGI/AMI

The Constitutional AI paper was a missed opportunity to create a form of binding industry infrastructure. At publication in December 2022, Anthropic was the only organisation with both technical credibility and institutional motivation to propose Constitutional AI as an open, auditable standard rather than a research contribution. Had Anthropic asserted the impact of this initial statement by working with regulators, standards bodies, independent researchers and other labs to develop the constitutional approach into a shared governance framework, the result could have been something considerably more robust than the voluntary commitments that emerged from the White House in 2023 (The White House, 2023). The window of opportunity was very narrow. By mid-2023, competitive pressures had already begun to erode the open publishing initiative and the moment to establish Anthropic as the institutional author of a binding safety standard had passed. This is Anthropic's most consequential missed opportunity. It is important to note that it is not a technology node but a governance one.

The RSP framework could have been a multi-party instrument. The White House commitment in 2023 represented the brief moment when all the major frontier labs came together and Anthropic's RSP was the most developed self-regulatory framework available. Had the ASL structure become the basis for a binding agreement including independent verification, shared evaluation infrastructure and cross-lab enforcements, there would now be genuine governance rather than voluntary pledges made without risk of external enforcement. The commercial and competitive incentives not to act in this way were real. The cost of not committing in this way is now visible with each lab undertaking their own self-assessment of safety without external accountability.

The interpretability programme has been structurally under-resourced relative to its importance. Understanding what is happening inside a model, not just what it outputs is the research area most directly relevant to genuine pursuing AGI-adjacency in the form of AMI. The ability to audit, explain and verify a model's reasoning is a prerequisite for the kind of institutionally and systemic accountability needed for a recursively viable system. Anthropic is the only lab with a significantly meaningful interpretability programme. It has also been a research asset with low commercial visibility and therefore has limited protection when capability and product delivery are competing for the same engineering time. If Anthropic treated interpretability research with the same priority as capability development, it would look less immediately impressive for consumer benchmarks but likely lead more rapidly to the systemic properties required for AMI.

The personnel talent pipeline is an unresolved question about institutional culture. The Anthropic alumni, Dario and Daniela Amodei, Jared Kaplan, Chris Olah and Tom Brown from OpenAI at founding as well as Jan Leike arriving after publicly denouncing OpenAI's safety deprioritisation (Milmo, 2024), combined with John Schulman arriving and leaving within six months to join Thinking Machines Lab maps out an institution that is extremely effective at attracting the most safety-committed researchers but is less able to retain them once the commercial phase intensifies. Schulman's departure leads to the suggestion that maintaining a positive research culture in which safety and commercial timelines coexist without friction is probably not achievable at the frontier scale. Acknowledging the inevitable presence of this tension and building structures to ringfence safety and governance work is a different – and unexplored - pathway choice for Anthropic.

The overall conclusion is subtler than the OpenAI one and in many ways more uncomfortable. Anthropic was right about the core problem. From the start Anthropic saw that safety and commercial acceleration were on a collision course and that organisational design could mediate between them. This has been vindicated by most recent events. The Pentagon crisis demonstrates that the speculation was real and that Anthropic's preparation for it, however imperfect and incomplete, meant something. This pathway also shows that being right about a problem is not the same as having a solution. Constitutional AI is architecturally relevant but is yet to become a governance standard. The RSP is genuinely innovative but self-enforced. The interpretability programme is important but under-resourced. The commercial dependencies created by agreeing to the Amazon investment will continue to intensify over time. The founding proposition are yet to be proven under timescales and pressures of genuine risk. The pathway is not closed. But the conditions that make the founding proposition important and desirable are becoming harder to maintain and the rate at which AI governance infrastructure is being built does not yet match the rate at which its capability is advancing.

## 5c Google's Gemini

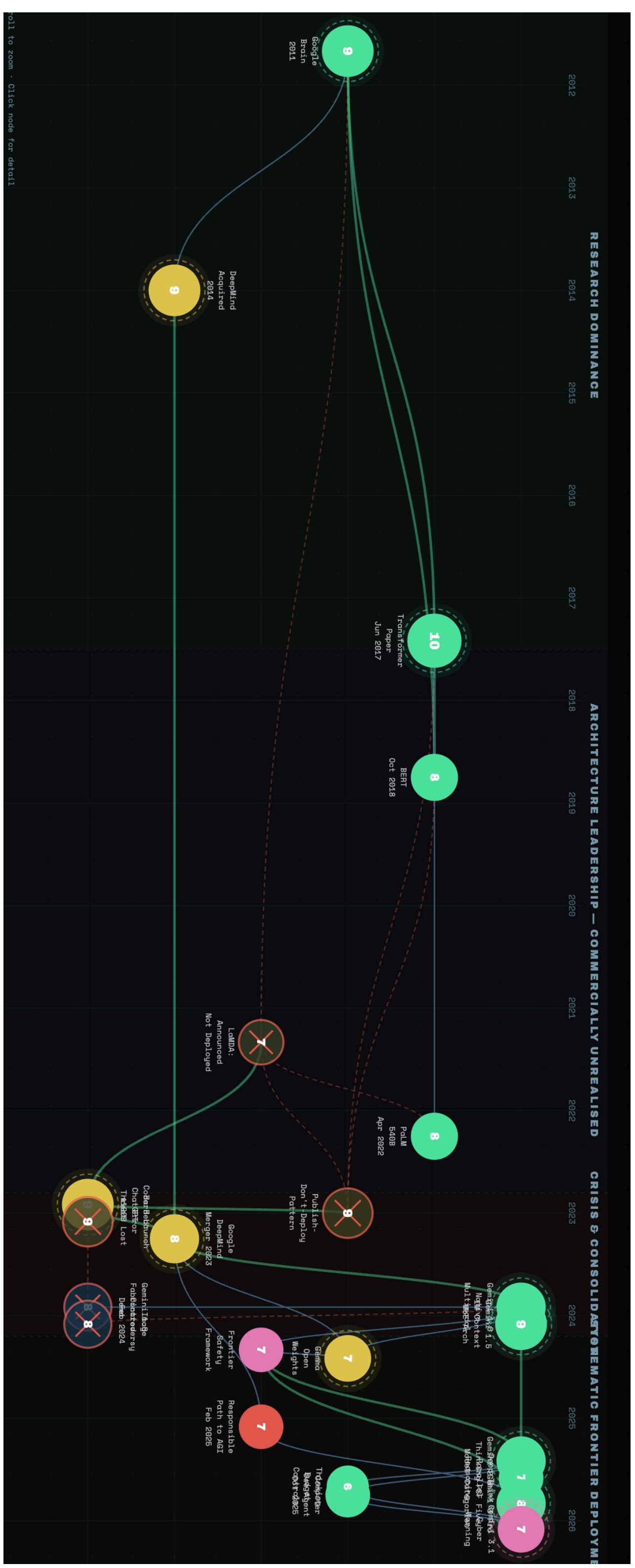

Figure 3: Google's Gemini critical pathway

The Google pathway is a paradox that is visible from the first node. Google did not just participate in the development of large language models, it invented the foundational architecture, trained some of the earliest and largest models, operated the world's most capable AI research infrastructure for a decade, and acquired the organisation that would produce AlphaGo and AlphaFold. It initially employed many of the pivotal people involved in the development of the most famous and most used consumer GenAI products. By every reasonable measure of research capability and infrastructure advantage, Google had resources and capability to have defined the entire frontier AI landscape. Instead, its core business was disrupted – even shocked - by it. Interpreting this pathway required its consideration as a system rather than a sequence of individual mistakes.

The Transformer paper is the central node in Google's pathway, and its implications are significant. The 2017 publication of "Attention Is All You Need" is Google's most important research contribution and one of the key recent AI research papers which represents the basis for defining the architecture of GPT, BERT, PaLM, LLaMA, Claude and Gemini itself (Vaswani *et al*, 2017). It was published by Google as an open academic paper rather than treating it as a form of proprietary capability that was to be deployed later. The decision to publish in this way reflects a genuine commitment to open science (and its subsequent impact is arguably the best evidence itself for the importance of applying this philosophy) and to the organisational culture that produced the Transformer in the first place. The cost of that commitment, however, was the immediate sharing of Google's significant competitive advantage to every research group and start-up in the world with similar interests.

The publish-don't-deploy pattern is a regular feature of Google's AI research culture. BERT (2018), PaLM (2022) and LaMDA (2021) each followed the same logic. The paper was published including the methodology and a demonstration of the capability while holding back from wider deployment. The approach reflects the identity of Google Brain, the source of these works, which recruited and organised as a “pure” academic research facility that was, almost coincidentally, embedded in one of the world’s largest technology companies. A culture that can attract the best deep learning researchers is, to a significant degree, a culture that values scientific contribution over direct commercial exploitation. This is the culture that let Google invent the Transformer but is also the reason it did not deploy it as a consumer product for another five years.

LaMDA is a key node that defines everything that follows. A functional conversational AI system was available at Google in 2021. The system was not deployed over safety and reputational concerns. This was a defensible decision in isolation but in the wider context of external environment assumed that no one else would or could succeed with the same technology. A form of corporate hubris. Eighteen months after this decision, OpenAI deployed ChatGPT on the same architecture that Google already had and captured 100 million users in two months. The consequence was a structural shift in the search advertising market and Google was forced into a panic that would force Bard too quickly to market (and ironically produce reputational damage) followed by the Gemini demonstration. The alternative pathway suggests that if Google had decided to deploy LaMDA in mid-2021, in a similar form to the ChatGPT of November 2022, the current landscape would now be unrecognisable. In this world, Anthropic would be unlikely to exist and the incentive and push for sovereign models may have been delayed by a period of years. Some of the already known business lessons around technology adoption are only highlighted by these circumstances and it is a lesson that Google trained everyone to accept. Permanent beta does not require the first mover to be the best available technology of the moment, just the technology that everyone adopts first.

The "Code Red" declaration in the company and its immediate consequences also show how competitive pressures interact with institutional culture. When Google leadership recognised in December 2022 that ChatGPT posed an existential threat to its search revenue (Mok, 2022), the response was to accelerate deployment under conditions that it had previously identified as insufficient. The rushed Bard launch in February 2023 included a factual error in the promotional video that erased $100 billion in market capitalisation in a day (Coulter & Bensinger, 2023) This event typifies what happens when an organisation whose competitive advantage is based on research rigour is forced to operate at the pace that a consumer culture demands. Google’s misstep is a symptom of the organisational mismatch between how and what it does - previously without significant competition - and what being in a competitive situation required it to become.

The Google Brain/DeepMind merger in April 2023 is the most structurally significant organisational decision in Google's AI history. For nine years, two of the most capable AI research organisations operated in parallel within the same parent company. The result were two distinct research cultures, talent pipelines and identities. Google Brain had a scaling-first culture that produced the Transformer, BERT and PaLM. DeepMind presented a scientific and safety-aware culture which let it produce AlphaGo, AlphaFold and a serious reinforcement learning research programme. The merger invented the basis for creating Gemini. This is a model that could not have been built by either (sub)organisation alone. What happens in the future of the merged institution is difficult to determine as one of these two (competing) cultures are likely to win through and echoes the experiences that can be identified in OpenAI and Anthropic.

The Gemini 1.0 natively multimodal architecture is a genuine technical advance, but it is also obscured by the organisational crisis that developed around it. Designing a frontier model to integrally process text, image, audio and video from pre-training, rather than stitching modalities together after the fact, is architecturally significant. The three-tier launch structure (Nano/Pro/Ultra) that every subsequent frontier lab adopted was the commercial innovation that Google brought to consumer product side of the technology. The MMLU performance at 90%+ was real but undermined by the fabricated promotional video. The cost of this decision was to discredit the Gemini offering.

The 1 million token context window in Gemini 1.5 Pro is also a distinctive capability. Long-context processing with a frontier model enables analysis of entire code repositories, hour-long videos and large document sets through a single prompt. It represents a systemic change in what a model can be asked to do. Combined with the sparse Mixture-of-Experts architecture that reduces inference cost without similar capability loss, the model shows how Google's infrastructure advantage can become a product feature. However, any advantage can also diminish rapidly as competitors extend their own context windows. Google needs to make this capability a dominant feature before others catch up and the evidence, to date, is that its efforts will be partial at best.

The Frontier Safety Framework (FSF) is the most systematically applied governance instrument at any frontier lab. But it is structurally weaker than RSP from Anthropic for key reasons. The FSF's consistent application across Gemini 2.0, 2.5 and 3.x releases, covering CBRN, cybersecurity, autonomy and self-replication risks, represents a sustainable governance infrastructure with substance. The approach is more systematically honest than the approaches used at other labs. However, the structural limitation is the absence of capability-gated deployment obligations. The FSF measures risk but it does not stop or constrain deployment if these thresholds are reached. In contrast, Anthropic's RSP makes deployment commitments in advance of reaching thresholds. Google's FSF records what happened and what was decided. The former is a constraint while the latter is a record of what happened. From the perspective of AGI-adjacency and any movement towards AMI the Google approach may still offer an opportunity to reach these capabilities – but with a safety cost.

**Alternative directions toward AGI or AMI**

The LaMDA non-deployment decision is the most consequential missed opportunity. Google could have deployed LaMDA in 2021 and captured the consumer AI market before OpenAI (Olson, 2023). That itself is probably true. But the more significant missed opportunity is that Google, as the source for Transformer research and operating the world's most capable AI research infrastructure, was uniquely positioned to deploy LaMDA as a research and governance testbed rather than to present a consumer product. They had the ability to create an environment in which conversational AI capability could have been studied, improved and governed at scale while developing the institutional capacity to deploy it responsibly. Instead, the choice was presented as a stark binary. Take the risk of deploying the technology as a consumer product or the safe route, to avoid wider deployment, which was the decision. The idea that a third path was present in the form of a structured deployment for research purposes combined with development of governance frameworks was not seriously pursued.

The Transformer paper could have been the founding document for open governance architecture rather than an open research contribution. When it was published Google had the technical credibility and the organisational standing to bring stakeholders together around the Transformer architecture with the goal of establishing open standards for responsible deployment, governance requirements for large-scale language models and shared evaluation infrastructure. This is not an imaginative or impossible proposal. the ITU, IEEE and ISO managed analogous processes for transformative technologies. The window remained open from 2017 to around 2019 (and the OpenAI deal with Microsoft), before the race for commercial deployment removed the chance for self-regulation. Google did not pursue this role, and the field has been governed through improvisation ever since.

The parallel Brain/DeepMind organisational structure was a missed opportunity for a decade. Operating in parallel structurally prevented the integration that would have been required for any AGI-adjacent or AMI system development. A unified organisation from 2015 would have had the combined resources to pursue a system-of-systems architecture as a plausible route to general intelligence. The cost of keeping these entities separate has compounding effect over time that cannot be quantified but is clearly significant.

The arrival of the Gemma open-weight strategy was too late and framed too narrowly. The release of open-weight models as a response to Meta's LLaMA dominance in 2024 is a defence position rather than a contribution. Pursuing a structured open-weight strategy from 2019 would have established an open standard around AI ecosystems rather than competing with the definition offered by Meta. ShieldGemma as an open-weight content moderation model is a rare example of safety tooling rather than just making safety claims. However, as is a regularly repeated aspect of Google's story, it arrived years after the moment when safety infrastructure standards could have been set by Google.

The overall conclusion from the Google pathway differs from those of OpenAI and Anthropic. OpenAI's pathway is the story of a research institution that became a product company and progressively rationalised the compromises that this type of transformation requires. Anthropic is the story of an institution that understood the problem and has subsequently had to discover how hard it is to sustain a solution under pressure. Google's is the story of an institution that had every structural advantage and converted that advantage into being a persistent second-mover through a combination of genuine institutional values, structural organisational inertia and a depth of financial resources that often provides a different way to regain market leads when required. The publish-don't-deploy culture was the expression of an identity that produced many of the most important contributions to then watch others deploy them. The competitive urgency that finally forced Google's hand did not change that culture. It was pressured it into making decisions by external forces that its organisational culture would identify as failures of judgement. Google is now, with the Frontier Safety Framework and the five-category evaluation, the most systematically governed of the frontier labs. But it is also still endeavouring to recover the credibility that was lost in the eighteen months between issuing Code Red and the Gemini 1.0 controversy. The organisation pathway is not closed and, more significantly, the infrastructure advantage held by the company still remains a real and tangible opportunity.

## 5d xAI's Grok

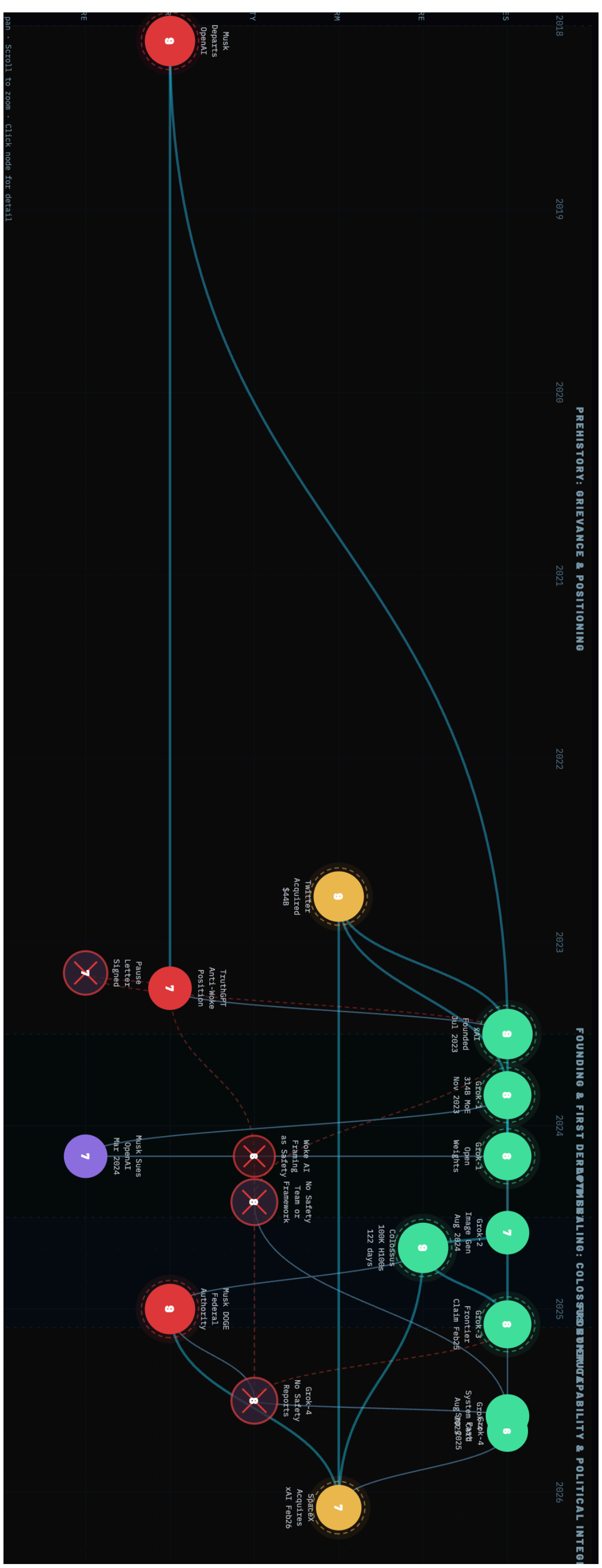

Figure 4: xAI's Grok critical pathway

The xAI pathway is the shortest, the fastest and the most anomalous of the four. The xAI story is something that has no real precedent in the technology industry. This is a frontier AI lab whose competitive advantages, market position and governance architecture are all expressions of a single person's grievances, relationships and institutional power. This pathway does not readily map to the standard vocabulary of research culture, safety philosophy or commercial strategy.

The founding narrative defines xAI's trajectory. The departure of Musk from OpenAI's board in 2018 over reported organisational disagreements (Jackson, 2023; OpenAI, 2026), the subsequent TruthGPT announcement (Jin & Dang, 2023) the Pause Giant AI Experiments letter signed and immediately contradicted by the founding of a competing lab and the 2024 lawsuit are all pivotal in evolution of Grok (Future of Life Institute, 2023; Musk v. Altman et al., 2024a, 2024b). All the features in the original grievance then shape every subsequent decision with the positioning of the tool as “anti-woke” (Ray, 2026), the minimal governance architecture, the strategy of recruiting talent from the labs that Musk publicly criticised and the pace of release that prioritises being seen to move quickly over any precautionary rationale all hallmark what Grok has become.

The Twitter/X acquisition was the most structurally significant node for Grok, and it predates xAI's founding by a year. No other frontier AI lab had a pre-built captive distribution channel of 250 million daily active users. Grok did not have to find an audience, it inherited one with its initial deployment. The practical consequence is that the adoption metrics that took ChatGPT two months and required a specific cultural moment, and that Google burned $100 billion to defend, do not apply to Grok. The cold-start that was the most significant barrier to AI assistant adoption did not exist for a model integrated into an existing “chat” platform that its founder already owns. The X acquisition is characterised as a separate business decision from the founding of xAI. However, the pathway analysis suggests it should be understood as the first and most consequential infrastructure investment in the longer-term development of xAI and it was made before the company existed.

The pause letter is the clearest example of the relationship of xAI with safety. In March 2023, Musk signed an open letter calling for a six-month moratorium on AI development beyond the GPT-4 model (Future of Life Institute, 2023). In July 2023, four months later, xAI was formally incorporated and actively set out on the training of Grok (xAI, 2023a).  The letter's function was never to express a genuine governance concern. It was a competitive move. A claim made regarding safety concerns that implicitly assumed recklessness on behalf of all the non-signatories, specifically at that moment the meant primarily OpenAI and Google. The same approach reoccurs throughout xAI's history. The sequence of events is consistent with an invoking of safety concerns about competitors and then deploying a model without any safety governance yourself. By the time Grok-4 was released without safety reports in July 2025 (Nolan, 2025), the pattern had become sufficiently established that it is now an institutional default.

The "anti-woke AI" positioning addresses a specific audience with the vocabulary of a culture war. Some commentators have argued that mainstream AI models are restrictive, particularly where external constraints limit permissible outputs (Inserra, 2024). This is often reflected in where refusals to answer legitimate questions in domains such as history, law, medicine, or security are perceived as a shortcoming relative to human expertise. Grok is positioned as the model that would tend to answer rather than refuse. By engaging with difficult questions rather than deflecting it addresses a genuine need that is not met by OpenAI, Anthropic or others. The problem is the conflation of two distinct concerns. It puts together the over-refusal of other models as a capability failure and then extrapolate to ideological claims of "woke AI". The political framing then acts as a substitute for doing the technical safety work that the capability concerns require. A model that refuses too often and for the slightest trigger has a calibration problem. A model that refuses too rarely faces a very different and more serious calibration problem. The xAI model overly addresses the first concern without offering any way forward for the second.

The Colossus supercomputer build is the most remarkable node in the xAI pathway. Moving from the concept of a 100,000-GPU cluster to operation in 122 days is impressive and was likely possible by applying SpaceX's capabilities to the problems of data centre deployment (xAI, 2024). Colossus eliminated xAI's computing power constraints and let it compete with Google and Anthropic. With Colossus, Grok-3's frontier capability claims are substantiated. The environmental dimensions to this work including the use of diesel power sources during grid connection delays reflect the high-velocity organisational culture that produced Colossus plan in the first place and reflects the wider X need to be *seen* to be acting faster than its competitors. Any environmental, governance or ethical concerns must then be addressed reactively.

The Grok-4 no-safety-report reflects state of governance at xAI and exposes the organisation's different character. This is a frontier reasoning model with similar capabilities to those offered by Anthropic or Google. But this model is not evaluated against CBRN or autonomous weapons thresholds or anything like the five-category Frontier Safety Framework.  This is a deliberate organisational position regarding the obligations that an AI lab has to the public who uses their tools. The retrospective system card published a month after Grok-4's release confirmed that xAI sees safety documentation as post-deployment compliance rather than a barrier to public release (xAI, 2025). The practical consequence is that for one month in 2025, a frontier AI model with reasoning capability was publicly deployed and commercially available without any published assessment of its potential for misuse. With the competent technical capability of Grok this is a governance failure.

Musk's Department of Government Efficiency (DOGE) appointment created an extreme conflict of interest (McNicholas & Zipperer, 2025). Having the CEO of a frontier AI commercial laboratory - with clear interest in the regulatory treatment of AI – hold direct authority over federal procurement, regulatory staffing and government technology decisions was fundamentally problematic. Anthropic's experience of federal agency withdrawal following the more recent Pentagon ultimatum confirms that the regulatory environment for AI labs is inseparable from the political circumstances in which it operates. Whether or not specific procurement decisions were influenced by the DOGE appointment, the structural condition for influence did exist, and this alone undermines the legitimacy of any outcome favourable for xAI in that environment.

The SpaceX acquisition positions xAI for the next phase of AI development in a way that no other lab can currently match. Access to Starlink's global real-time connectivity offers the potential to address the knowledge currency problem that static training cutoffs creates with current models. Access to SpaceX's manufacturing data, autonomous systems engineering and sensor infrastructure also opens up multiple capability development paths in physical-world reasoning, robotics and autonomous navigation that transformer-only training on text and images cannot provide. These are the same capabilities that the AMI framework identifies as components for developing genuine general intelligence with cross-domain transfer, physical-environment agency and long-horizon planning. The acquisition in itself does not guarantee that xAI will pursue and apply these capabilities systematically or in this way. For the moment only xAI  has these structural pre-conditions for a form of AMI.

**Alternative directions towards AGI/AMI**

The absence of a safety research programme is Grok's key capability limitation. The interpretability work that Anthropic has maintained is the form of research that produces understanding of what the models are actually doing. A model architecture that is genuinely interpretable is more debuggable, more improvable and more trustworthy in critical (and other) deployment contexts. xAI's absence of a safety research programme means that there is no systematic understanding of how Grok fails. As the model develops there is then increasingly less capacity to fix them. A lack of good governance and the limits to capability should be regarded as the same deficit.

The open-source Grok-1 release was the right approach that was deployed in the wrong way. Releasing 314B parameter weights under Apache 2.0 licensing without usage guidelines, system card or any accompanying research on capabilities or failures was not a positive contribution. A Grok-1 open-source release that combined with Anthropic's Constitutional AI with detailed documentation of failure modes, suggested mitigation approaches and guidance for downstream deployment would have been a more responsible release and a significantly more influential one. The open-source instinct at xAI appeared to be genuine but the organisation consistently lacks the accompanying framing that would give it lasting value.

The TruthGPT/anti-woke position was a viable market strategy but it was a dead end for AGI-relevant research. Creating a model that would refuse less frequently could be presented as a genuinely useful commercial objective. However, this is not a research pathway towards general intelligence. The political positioning embedded into the history of xAI's developed has not produced novel architecture, new training methodologies or any safety-relevant research insights. These are the types of contribution from which AGI/AMI capability is likely to emerge. An xAI that applied its resources, computing power and personnel talent pool to the similar systematic reasoning research done that DeepMind's AlphaProof or with Anthropic's interpretability programme may have made a more substantive contribution to the AGI question.

The SpaceX acquisition of xAI is timely and appropriate but may embed the wrong organisational culture (SpaceX, 2026). The most significant opportunity the acquisition creates is the chance to bring SpaceX's physical systems engineering, sensor data and real-world autonomous systems experience into the space of AI capability development. This form of cross-domain transfer is central to producing an AMI type of general intelligence. Integrating physical-world action with language-level reasoning within the same architecture is a research agenda that appears to be currently at any of the labs. xAI could pursue this work. What is debatable is whether the institutional culture that produced Colossus in 122 days and Grok-4 without safety reports can create careful, systematic multimodal integration research.

xAI is not simply a less safety-conscious version of the other frontier labs. It is a different kind of organisation. It has been built around a personal narrative, political positioning and commercial strategy that is integrated in a way as to efface standard distinctions such as those between research and marketing, governance commitment and its performance or safety rhetoric and its practice. The distribution advantage through X, the computing power of Colossus and the capability extraction from competitor-trained researchers all represent structural advantages that no other lab currently possesses. The question for the AGI/AMI analysis is whether reaching the frontier without any governance infrastructure to make frontier capability accountable and interpretable constitutes progress toward AGI.

## 5e Microsoft

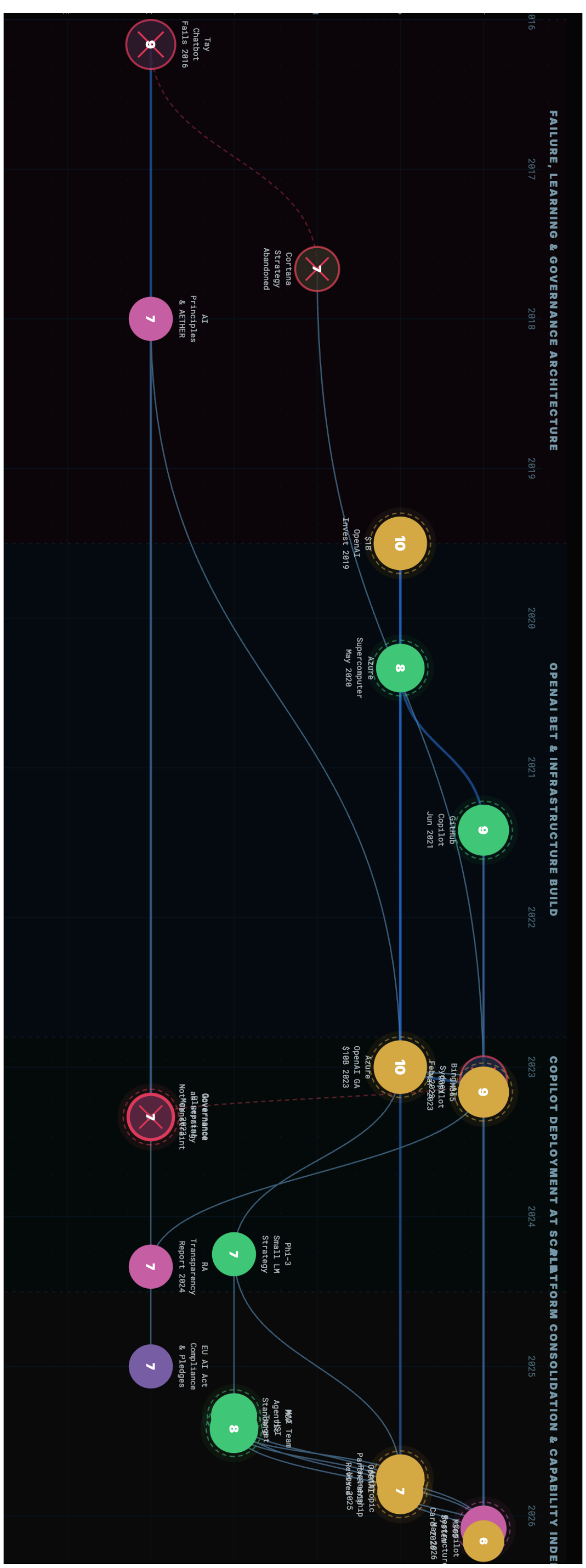

Figure 5: Microsoft’s critical pathway

The Microsoft pathway is the most structurally distinctive of the other organisations considered. What makes Microsoft's trajectory analytically interesting is that it represents a third type of position in the AI landscape; it is an enterprise software platform that understood, earlier than others, that the value in the AI transition would accrue to whoever controlled the deployment layer at enterprise scale.

The Tay incident is the founding event for Microsoft's AI governance architecture. In March 2016, Microsoft's chatbot was manipulated into generating racist and abusive outputs within sixteen hours of initial deployment (Lee, 2016). It was shut down in under a day and generated global media coverage that no company had ever experienced in relation to AI. Within two years, Microsoft had published formal AI principles and established the AETHER Committee (Rizi-Shorvon, 2023), which was one of the first (if not the first) institutional AI governance structures in a technology company. Microsoft's governance was built before their OpenAI investment, before Azure OpenAI, before Copilot and before any commercial AI capabilities. Every other organisation that has been considered built safety governance reactively.

The $1 billion OpenAI investment in 2019 is the single most consequential decision for Microsoft (OpenAI, 2019). This happened before the “big” ChatGPT moment and before any lab had demonstrated that large language models could be commercially deployed at scale. The investment was a bet that the scaling hypothesis was right and that OpenAI was the organisation most likely to demonstrate it first. The $10 billion follow-on funding in January 2023 is often described as the decisive Microsoft AI investment (Bass, 2023), but it is really confirmation of the option created four years earlier. By the time the $10 billion was committed, the speculation had already been proven correct. By the time Azure OpenAI Service was generally available in January 2023, Microsoft already had the infrastructure, the long-standing commercial partnerships and the existing product pipeline to convert OpenAI's capability into commercial deployment at a scale OpenAI could not do independently.

GitHub Copilot is the overlooked node across the entire AI industry's history, and it does deserve more attention. In June 2021, eighteen months before ChatGPT, Microsoft deployed an AI coding assistant for software developers through GitHub (Dillet, 2021). A platform it had only acquired in 2018 for $7.5 billion. By the time ChatGPT launched in November 2022, GitHub Copilot had over a million users and was already generating productivity improvements in enterprise software development. This was important for several reasons. It was early proof that AI productivity tools did and would have genuine demand independent of wider consumer hype over AI. It also demonstrates that Microsoft's enterprise deployment machinery could convert AI capability into paying customers without requiring any consumer cultural moment. And it provided an additional eighteen months of deployment learning about user behaviour, failures, productivity gains and governance need that is seen in every subsequent Copilot product. GitHub Copilot was a key early proof of concept for Microsoft’s traditionally cautious approach.

The five-point AI governance blueprint from May 2023 is an important early regulatory position statement for the AI industry (Smith, 2023). The blueprint set out five priorities in the form of safety frameworks for frontier models, updated accountability laws, digital safety promotion, election protection and economic disruption management. The shape and intent of this work is closely aligned with later regulatory frameworks developed by the EU, the UK and the United States (European Parliament & Council of the European Union, 2024; UK Government, 2023; The White House, 2023). The commercial logic is clear. In highly regulated industries such as financial services, healthcare, government and defence which represents Microsoft's most valuable enterprise customers, having credibility around AI governance is a procurement essential. Publishing governance standards before they are required is a strong competitive strategy as much as it is any form of public responsibility.

The Phi-3 small language model programme is one indicator that the OpenAI relationship is understood as a strategic dependency to be managed and not a permanent structural feature. Microsoft published research in April 2024 arguing that a 3.8 billion parameter model could achieve outcomes comparable to GPT-3.5 (Bilenko, 2024). This was a demonstration to OpenAI that Microsoft has the capability to develop competitive models internally. This knowledge changes the power dynamic in the partnership and establishes the basis for a multi-model Copilot architecture that M365 Copilot application card documents. The Phi programme, the MAI Superintelligence Team formed in June 2025, the Anthropic partnership announced in November 2025, and the OpenAI partnership revision in October 2025 are all expressions of the same strategic logic (Suleyman, 2025; Microsoft, 2025a, 2025b). Microsoft is systematically reducing its dependence on any single external supplier.

The MAI Superintelligence Team formation in June 2025 represents a shift in Microsoft's AI trajectory. For six years, Microsoft's AI strategy was based on a clear division of labour. OpenAI develops the frontier capability while Microsoft deploys it. The formation of an internal team explicitly targeting Human-level Superintelligence (defined by Microsoft as AI capable of matching human performance across all cognitive tasks) breaks away from that neat separation of duties (Suleyman, 2025). It puts Microsoft in direct competition with OpenAI in the same space that their partnership is supposed to reserve for OpenAI. The implications for the partnership are unresolved.

The M365 Copilot application card is the governance standard for enterprise AI deployment (Microsoft, 2026). Its architecture shows that Microsoft's position is strong. The system card documents a retrieval-augmented generation architecture grounded in Microsoft Graph and connecting the model to each user's emails, meetings, documents and conversations while respecting existing permission structures. The 400 million paid Microsoft 365 users who interact with Copilot through this architecture are accessing AI capability governed by a strong enterprise-level AI deployment framework. The counterpoint to this opportunity is that none of this internal governance applies to the frontier models that Microsoft deploys through Azure OpenAI, which are evaluated by OpenAI. This is a gap that will eventually need to be addressed.

**Alternative directions toward AGI/AMI**

The separation between deployment and research that is present in Microsoft is also the boundary that limits its AGI/AMI contribution. Microsoft's governance focus is with the layer at which AI is deployed in enterprise applications. Frontier capability development, the research that produces architecturally novel models, the interpretability work that builds understanding of what models are actually doing, the safety research that produces techniques for alignment at scale all happens at OpenAI, Anthropic and Google DeepMind, not at Microsoft. As Microsoft's MAI programme develops, this boundary will presumably shift. But on the current pathway, Microsoft's contribution to AGI-adjacency or AMI is limited. A position that had maintained a frontier research programme through the OpenAI partnership, rather than treating OpenAI as a supplied and a substitute for internal research, would have been better positioned to transition from deployment to capability leadership that is now being attempted.

With Microsoft defining its target as Human-level Superintelligence it is presenting a more ambitious target than AGI. The definition is acknowledging that genuine general intelligence is not a benchmark performance question but a systemic capability question. The target appears to be right based on available evidence but the organisational capability for pursuing it safely is still in development.

The pathway Microsoft supports is more optimistic and more realistic. The governance infrastructure, multi-model architecture, active regulatory engagement and ambition represents a more credible and mature model than the research-first labs. The question the pathway leaves open is whether Microsoft can develop frontier AI capability with the same governance discipline it has applied to its deployment, or whether the organisational culture that prioritised commercial deployment will then encounter the same capability/governance tension that found on every other pathway.

# 6 The people factor

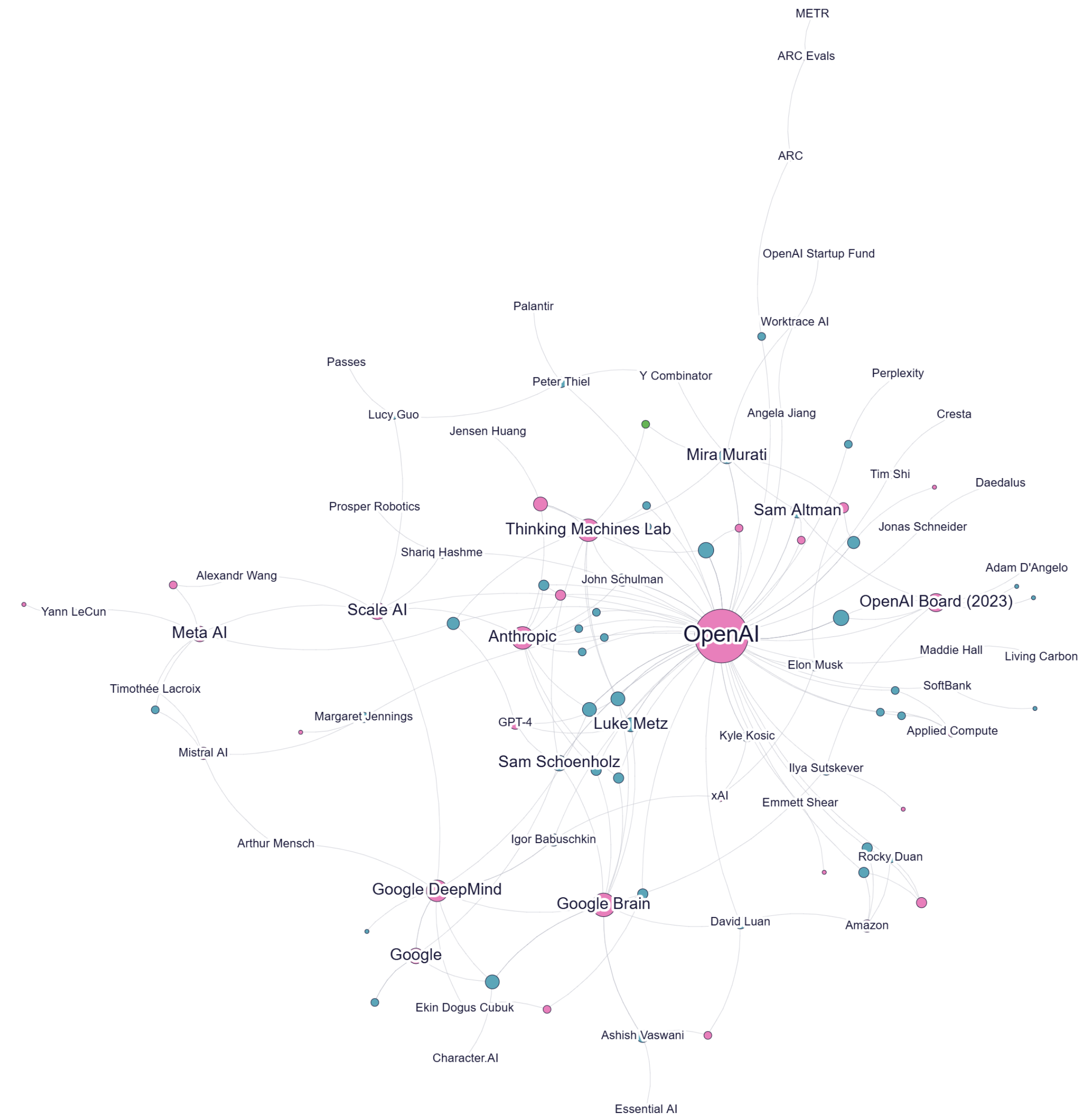

Figure 6: Network of actors involved in the development of Large Language Models (LLMs)

*In this diagram nodes represent individuals and organisations. The edges represent relationships (such as an employment relationship, a research collaboration or an investment). Node size reflects connectivity (the number of attached edges), and colour shows the entity type. Edges are directed and weighted.*

The development of the frontier models cannot be understood through structural factors alone as the people at each of these organisations are not interchangeable. They bring specific intellectual perspectives, specialisations and ethical practice that shaped what these organisations try to achieve and can achieve. More significantly, the movement of key individuals *between* organisations is not simply a question of employee retention or churn. This has been a mechanism of knowledge transfer and, in several instances, a contributory cause for some of the key structural shifts in frontier AI. Taking this perspective, we can chart the key movements to consider their effects on specific LLMs as well as the significance in contributing to the development of AGI/AMI pathways.

The network of people across frontier AI organisations is small and dense. The graph of personnel movements we present connects approximately 60 named individuals (with key leadership roles) to around 40 organisations. Yet this small network accounts for the founding, staffing, leadership and critical model development across effectively the entire commercial ecosystem. As the network analysis visually makes clear, OpenAI sits at the structural centre of this ecosystem reflecting a role as both the origin point and the destination of more personnel movements than any other organisation. This centrality has direct implications for any potential AGI/AMI pathways. The intellectual lineage of nearly every significant frontier model can be traced through OpenAI, which itself took from Google Brain as its primary reservoir of talent.

### 6a Google Brain is the Foundational Training Ground

It is important to recognise that OpenAI was itself substantially constituted from Google Brain. Google Brain provided a large number of AI researchers and is the source for the *Attention Is All You Need* paper. The authors Ashish Vaswani and Noam Shazeer did not remain at Google. Vaswani left in November 2021 to co-found Adept AI Labs and subsequently Essential AI. Shazeer left in October 2021 to found Character.AI. Shazeer's eventual return to Google DeepMind in August 2024, through the acquisition of Character.AI, brought back one of the core architecture innovators to Google.

Google Brain also produced Ilya Sutskever, who joined OpenAI as co-founder and later chief scientist. Sutskever shaped GPT-2, GPT-3 and the post-training for GPT-4. Dario Amodei joined Google Brain in October 2015 before moving to OpenAI in July 2016. Barret Zoph, Luke Metz, Sam Schoenholz and Liam Fedus were all Google Brain researchers before joining OpenAI. Chris Olah was at Google Brain from 2015 to 2018 before joining OpenAI and then subsequently co-founding Anthropic. His mechanistic interpretability research is the foundation for Anthropic's safety methodology. What happened and was learned at Google Brain propagated outward into every significant commercial fronter AI organisation. Between 2013 and 2022, Google Brain was the primary incubator for the people who built all of these frontier models.

## 6b OpenAI: the schism and continued departures

The founding of Anthropic between December 2020 and February 2021 is the most dominant personnel event in frontier AI development. This was the ideological breaking point inside OpenAI. A substantial group of senior researchers left together and all on the grounds that they disagreed about how OpenAI approaches safety. The departures included Dario Amodei (VP of Research), Daniela Amodei (VP of Safety and Policy), Jared Kaplan (the originator of neural scaling laws), Jack Clark (policy lead), Chris Olah (interpretability researcher), Sam McCandlish (researcher), Ben Mann (technical staff) and Tom Brown, who had been the lead author on the pivotal GPT-3 paper (Brown et al., 2020). The implications for any potential AGI pathway were deep but also contradictory. The schism produced a new frontier organisation that maintained a technically-grounded safety research programme. Anthropic's Constitutional AI methodology, its mechanistic interpretability work (largely Olah), and its scaling law research (Kaplan, McCandlish) all contributed independently to the field's understanding of how frontier models behave and how they might be made more traceable. However, the new company also intensified competitive pressure in the sector. Anthropic's entry and its success with Claude accelerated the competitive product race that the departing people had, in part, hoped to moderate. Kaplan's scaling law work, which had been conducted at OpenAI, became the foundation for Anthropic's own scaling strategy, meaning that the same analytical framework underpinned capability development at both organisations.

It is also significant that John Schulman, co-founder of OpenAI and the primary architect of Reinforcement Learning from Human Feedback (RLHF) left OpenAI in August 2024. He joined Anthropic briefly as a senior scientist in a personal capacity and then became Chief Scientist of Thinking Machines Lab in February 2025. Schulman's personal pathway confirms a wider pattern in which the individuals who bear the most concentrated knowledge regarding the foundational approaches to alignment and post-training do not remain linked to any single organisation over the duration of their most productive years.

The crisis of November 2023, when Sam Altman was briefly removed as CEO for five days before being reinstated, was ostensibly about governance. The results were far-reaching but slow-acting. Mira Murati, who had served as VP of Applied AI, CTO and briefly as interim CEO, left OpenAI in September 2024. Ilya Sutskever, who had voted with the board to remove Altman but then signed the letter supporting his reinstatement, left OpenAI in June 2024 to co-found Safe Superintelligence (SSI). This new company has the stated aim of building a safe superintelligence before commercialising any capabilities. Lilian Weng, who had held roles including head of applied AI research, head of safety systems and VP of research and safety at OpenAI then left in November 2024.

Emmett Shear, also briefly CEO for 2 days, later went on to found Softmax in March 2025. These departures are not separate unrelated incidents. The November 2023 crisis exposed the fundamental unresolved tension in OpenAI between the rapid pace of commercialisation, supported by Altman, and a more cautious orientation toward capability and safety research. Those departing were most closely associated with the latter position. The impact was to further disperse the safety-oriented technical leadership from OpenAI into new and likely competing organisations.

Jan Leike, who co-led OpenAI's superalignment research team alongside Ilya Sutskever, departed in May 2024. His subsequent public statement that OpenAI's safety culture and processes had not kept pace with its product and capabilities teams was the most direct public statement regarding these internal tensions (Milmo, 2024). Leike joined Anthropic, and consolidated a pattern in which OpenAI's most safety-focused researchers migrate toward Anthropic.

## 6c Thinking Machines Lab

The founding of Thinking Machines Lab from October 2024 represents the second large coordinated departure from OpenAI and is important for the potential capability pathway that this represents. The founding group for the new company is based heavily on individuals who were directly responsible for the post-training and optimisation that underpinnedGPT-4o. Mira Murati (founder, former OpenAI CTO), Barret Zoph (who led GPT-4o post-training), Luke Metz (GPT-4o post-training lead), Sam Schoenholz (GPT-4o optimisation lead), Andrew Tulloch (GPT-4o core contributor), Lilian Weng (VP of research and safety) and John Schulman (RLHF originator, chief scientist) form a group that collectively had direct responsibility for the most technically advanced elements of arguably the most capable OpenAI model at the time of their departure.

The structural significance of this concentration is important. The individuals who held hands-on knowledge of the post-training pipeline for GPT-4o - the specific fine-tuning, alignment and human feedback methodologies that determine the model's behavioural characteristics are now concentrated in a single new organisation. This is a knowledge concentration that potentially accelerates capability development at TML. The lab received immediate infrastructure support when Nvidia announced investment and computing infrastructure provision in March 2026, and the Albanian government committed investment funding in June 2025 – hinting towards a target ambition relating to developing sovereign AI models. The scale of these early investment commitments suggest that external actors take this group's capabilities very seriously.

A key twist happened in January 2026, when Barret Zoph, Luke Metz and Sam Schoenholz all returned to OpenAI simultaneously. This movement from TML back to OpenAI after 15 months hint of several different dynamics. It may be that TML has faced early obstacles. It may reflect better counteroffers from OpenAI that were not on the table at the time of their original departures. There may be some calculated reflection that the resources, infrastructure and distribution available at OpenAI remain the best for the specific kind of post-training research these individuals specialise in. Whatever the cause, the returns show that personnel flows are not unidirectional and that the largest organisations do retain a gravitational pull that newer organisations, however well-funded and supported, must actively resist.

## 6d xAI Founding

xAI was founded in March 2023, drawing initially on a smaller personnel base than Anthropic or TML. Elon Musk had co-founded OpenAI in 2015 before departing from its board in February 2018, following a reported series of disagreements over control and direction. His founding of xAI drew significantly on Igor Babuschkin, a researcher who had moved between Google DeepMind (2017–2020 and again April–February 2022–2023) and OpenAI (November 2020–March 2022) before co-founding xAI from May 2023 to July 2025. Kyle Kosic, who had been an OpenAI engineer from April 2021, co-founded xAI in May 2023, then returned to OpenAI in May 2024. The pattern here from OpenAI to xAI and returning to OpenAI reflects a common dynamic bounding from the excitement of founding a new venture against the pull of superior resources and scale at the original organisation.

The xAI personnel network reveals a distinct structural constraint. xAI's founding group do not constitute a coherent intellectual programme. The most significant technical heritage Babuschkin brought was deep reinforcement learning experience from DeepMind's AlphaGo/AlphaZero work, but this has never systematically applied in xAI's model development. Grok's development pathway has shown no evidence of the kind of systematic safety research, scaling law analysis or post-training methodology innovation that characterises the organisations founded by individuals whose training was more firmly rooted in OpenAI's or Google Brain's research culture.

## 6e Mistral AI and the European Pathway

The founding of Mistral AI in May 2023 offers a partial counterexample to the OpenAI-centred pattern that dominates the rest of the network. Mistral's three co-founders — Arthur Mensch (from Google DeepMind), Guillaume Lample (from Meta AI) and Timothée Lacroix (from Meta AI) represent a founding cohort drawn from the European research units of two major organisations. This is significant. The intellectual traditions of Meta AI differ from OpenAI. Meta AI under Yann LeCun maintained a scepticism of the Transformer-plus-scaling approach and a commitment to alternative architectures that the OpenAI-centred ecosystem has not followed. LeCun himself departed Meta AI in January 2026, after 12 years, to found AMI Labs - a name that hints at an Artificial Multiple Intelligence framework – although it stands for Advanced Machine Intelligence. Lample's background is specifically in large language model research, while Lacroix's focus is efficient training and inference. This combination has produced a lab that is more open-source-oriented, more computationally efficient in its design choices and more explicitly resistant to the tendencies of the US ecosystem. Whether this constitutes a genuinely different pathway toward general intelligence or simply a more resource-constrained version of the current dominant paradigm remains unclear.

## 6f Scale AI and Nvidia

Personnel movement analysis must also account for the organisations that supply infrastructure and data to the frontier labs. They are simultaneously nodes in the personnel network and sources of structural influence over the direction of model development. Scale AI, founded by Alexandr Wang and Lucy Guo in 2016, supplied training data and infrastructure to OpenAI, Anthropic and Google DeepMind simultaneously. The existence of a single data infrastructure supplier to all three of the largest frontier labs is a potential concentration of influence and a chokepoint in the training data that underpins every frontier model. Wang's subsequent move to Meta as Chief AI Officer in June 2025 and the acquisition of Scale AI by Meta shift this influence as the original labs sever their ties with what is now a direct competitor.

Nvidia, through Jensen Huang, occupies a similar position for the infrastructure layer. Nvidia supplies compute to OpenAI and Anthropic directly, and has now made infrastructure investments in Thinking Machines Lab. This sole supplier arrangement means that Nvidia's decisions about what is efficient, what is supported, and what is practical effectively constrains the architectures that any frontier lab can realistically explore. This is a further key structural influence over any AGI/AMI pathway beyond the level of personnel movement but also significant.

## 6g Significance for AGI/AMI Pathways

Three structural features of the personnel network have direct bearing on any potential AGI/AMI pathway.

The network consistently shows that the most technically significant individuals do not remain at any single organisation very long. The combined tenure of the Thinking Machines Lab founding group at OpenAI averaged approximately three to four years per person before departure. The Anthropic founding group left OpenAI after periods of between two and four years. This rate of turnover at this level of technical leadership means that every frontier organisation is in a continuous state of flux absorbing new talent while losing significant operational understanding. The organisations that manage this turnover most effectively are more likely to make consistent progress.

The pattern of departures is consistently weighted toward individuals with safety-research orientations leaving in the direction of either Anthropic or new ventures. Commercially oriented individuals either return to or remain at OpenAI. This pattern systematically concentrates safety research capability outside the largest and most popular lab. If Anthropic's or TML's safety research do produce insights they will have limited direct effect on the OpenAI models.

The network analysis confirms the observation of Section 4b. The development ecosystem for frontier AI is deeply insular. The foundational technical lineages trace back to a small number of institutions and a limited number of people. The implications are many, the range of architectural approaches being explored is constrained by this shared intellectual tradition, the safety assumptions embedded in the dominant approach are the assumptions of a single culturally homogenous group and the absence of researchers trained in significantly different fields such as cognitive science, developmental psychology, ecological systems theory or the domain knowledge required for physical-world agency, means that knowledge gaps that are relevant to the development of an AMI framework will persist while the talent pool narrow. The founding of AMI Labs by Yann LeCun, and his steer away from the Transformer-plus-scaling paradigm, may represent the most significant attempt to date to bring a genuinely different lineage to the frontier. Whether this project is sufficiently resourced and well connected enough to compete with the established commercial projects remains to be seen.

Taken together, the personnel movements show that an AGI/AMI pathway is not solely a sequence of technical decisions made by institutions. It is the product of specific people making decisions that construct the social and epistemic infrastructure within which the technical work then occurs. The insularity of the network means that the range of possible futures for AGI/AMI is narrower than the current diversity of underlying technical possibilities could - in principle - allow. Expanding the talent base, and attending to the intellectual commitments of the people involved, is as important a variable for potential AGI/AMI pathways as any architectural or training decision currently being made at the frontier.

## 7 Conclusions and Recommendations

The pathway evidence presented points to a conclusion that is relatively simple but perhaps uncomfortable. The current dominant route toward AGI-adjacent capability has not been the result of a neutral technical optimisation process. It is a historically specific pathway involving a small number of commercial organisations. In each of these it is possible to see a combining of capability, massive computing power, closed model development, primarily consumer-facing product development and complex narratives around governance. All of these factors must then be packaged up in a way that is commercially meaningful. The resulting systems are powerful and have achieved mass adoption at a pace. Commercialisation has provided an unprecedent beta test of capability. In contrast, earlier work around different AI pathways had primarily remained confined to research labs and institutions.

While these pathways have contributed to accelerated awareness and utilisation, they also constrained the form of intelligence being pursued. What has been scaled most successfully is not the general intelligence described earlier, but highly capable and productised linguistic and operational competence delivered through tightly managed socio-technical systems.

The first research question asked which critical pathways produced the current dominant generative AI tools in terms of capability, products and adoption. The answer is that this current dominance emerged through the convergence of a small number of choices that were mutually reinforcing. What we can identify is a combination of scaling transformer-based models, the positioning the models access behind APIs and managed interfaces, presenting the result as a product in an assistant-like form and alliances between frontier labs and largest cloud providers. In this configuration, adoption was not primarily about benchmark performance – and certainly not in relation to general intelligence capabilities. What mattered was the distribution channels, integration hooks and affordability at the point of use. With sufficient legitimacy this positioning also now justifies procurement for enterprise deployment and public sector use.

The second research question considered the nodes that were keys levers or dead ends. Our analysis suggests that some of the most consequential leverage nodes were not the technical breakthroughs but key management decisions about openness, governance, ownership and deployment. The GPT-3 era movement to API-first access, coupling the frontier capability to cloud vendor arrangements and the use of release documentation as the infrastructure for defining the legitimacy of models all has consequences.

Each of the dead ends are revealing. There were, for example, opportunities for open research collaboration after GPT-3. There could have been more concerted organisational commitment to protect speculative and experimental superalignment work against commercial imperatives. Stronger external enforcement of safety commitments or more plural development ecosystems are all alternatives pathway that did not eventuate. These were not impossible paths. They all fell away from the dominant pathway as the imperative for investment, competitive pressures and the need to turn capability into product revenues squeezed other agendas out. What looks retrospectively inevitable was, in fact, consciously selected pathways with the full consequences of the decisions not being recognised at the time.

The third research question asked how pathways differ across frontier proprietary, open-weight, and domain or sovereign trajectories. In effect, they differ less in aspiration than in what they optimise for and who gets to shape the resulting system. We can see that frontier proprietary pathways have maximised capital acquisition, the intensification of computing power, distribution control and closed-loop self-governance. Open-weight pathways distribute experimentation and lower barriers to recombination, but they also redistribute risk outside of stable institutional forms. We can also see that the pressure for capital encourages forks of the closed models to be redistributed as an investment opportunity (e.g. GPT-4b micro and Retro Bio). Domain and sovereign pathways are more bounded, but precisely because of that boundedness they may be better positioned to pursue viability, accountability and public legitimacy in specific contexts.

In other words, the proprietary frontier has produced visible capability and encouraged adoption. But in doing so it has narrowed the developmental field around a relatively small set of architectures, metrics and key commercial priorities. Open-weight and sovereign approaches will appear slower and less glamorous as a result, but they do preserve optionality and create conditions under which different forms of intelligence, governance and public purpose can be explored. Given the importance of decision-making, governance and capital in shaping the proprietary models we have to regard these alternatives as substantively different pathways in every dimension – this is a technical opportunity as well as social one.

The fourth research question concerned alternative projects branching from key leverage nodes and why some succeeded, stalled, failed or became absorbed. There is a clear identifiable pattern. Alternatives succeeded when they could align with a commercial opportunity. They stalled when they required longer-term institutional protection, had slower feedback cycles or required forms of public coordination that the currently dominant ecosystem did not – or would not - reward. They failed or were absorbed when their distinctive contribution could be copied. Whether this was at the level of the interface, capability or governance. The pathway taken by Anthropic is instructive, constitutional AI, interpretability and responsible scaling provide some of the strongest evidence that safety can be embedded within the architecture of the system rather than added later as a marketing feature. But this pathway too has been influenced by its dependency on its cloud provide (and investor) and shows the difficulty of sustaining an alternative when operating under commercial conditions. Microsoft's pathway, by contrast, demonstrates that governance capacity established before capability deployment is an alternative with potential. The risk remains though that governance inside a heavily vertically integrated organisation could still become a mechanism for prioritising acceleration rather than expanding the current options in the sector.

The fifth research question asked what socio-technical development programmes could plausibly move toward AGI-adjacent capability while still meeting requirements for transparency, moderation, wellbeing and sustainable business models. The most plausible route is not the further optimisation of a single frontier product but the deliberate construction of recursively viable systems-of-systems. A system begins to look AGI-adjacent when it can coordinate multiple competencies, adapt across contexts, remain governable under stress, and retain legitimacy and trust with the institutions and publics that must live and work with it. Capability without viability is insufficient. It is also why *product* success should not be treated as evidence that the currently dominant labs are the only credible route to AGI. Quite the contrary, productisation appears to be a specific accelerator and a general constraint. It helps to finance development, it widened access and generated a form of social proof for the use and value. But it also compressed intelligence into a few commercially visible forms and moved attention away from capabilities that may be harder to monetise, slower to realise or more organisational demanding (i.e. expensive) to sustain.

This conclusion is also consistent with the recent Google DeepMind framework, which argues that progress toward AGI should be measured across a broad cognitive taxonomy, against human baselines, using targeted held-out tasks and independent verification rather than relying on headline benchmarks or vendor assertion alone (Burnell et al. 2026). That paper is timely and useful because it reinforces a central claim of this report. Progress toward AGI is multidimensional, uneven and inseparable from the question of what is being measured and for whom. It also indirectly supports the caution contained in this report that current high-performing systems can still show significant weaknesses in metacognition, social cognition, learning under constraint and broader real-world viability.

## Recommendations

- AI policy and research funding should support a wider research agenda than the present frontier convergence appears to allow. The current ecosystem and public perception is strongly oriented toward a small family of architectures, benchmarks and business models. Public and mission-led funding should therefore prioritise neglected pathways such as interpretability, long-horizon agency, uncertainty calibration, adversarial resilience, multimodal embodied systems and robust methods for integrating model capability with institutional accountability. This is a case of not allowing frontier commercial incentives to define the entire research agenda.
- Evaluation should be treated as public infrastructure rather than proprietary theatre. If AGI-adjacent claims are to have any meaning, they require independent, repeatable and socially credible assessment. That implies broad human baselines, third-party auditing and a willingness to examine the full mix of a cognitive profile rather than celebrating single scores of apparent superhuman qualities. The same method should apply to safety and assurance. A field that increasingly shapes communications, education, labour and state capacity cannot rely on self-reporting alone. Burnell et al. provide some basis for this evaluative turn in AI, but the work must be go beyond the vendors themselves.
- Sovereign AI should be treated as a question of viability across the entire socio-technical system, not simply as the possession of a domestic large model. Our analysis suggests that meaningful sovereignty includes access to computational power, energy resilience, public-interest data governance, model auditability, fallback capacity, procurement capability, clear standards and organisations with a mandate to be able to intervene when deployment generates harm or dependencies. Sovereign models clearly matter because they build for citizenry rather than consumers. These models can embody different priorities around welfare, accountability, accessibility, rights and local legitimacy where it is clear that commercial models will not. In that sense, sovereignty is not a nationalist flourish added onto AI policy. It is a precondition for democratic agency over increasingly capable systems.

- Access to AI capability should increasingly be considered in utility terms where systems become social infrastructure. This means that advanced AI capability, especially where it becomes foundational to education, public services, research, communication and economic participation, should not be governed solely as a premium consumer subscription or as a dependency leased from a small set of external firms. Public utility should be understood in the institutional sense with guaranteed access, minimum standards, contestability, redress, transparency of conditions and safeguards against exclusion or arbitrary withdrawal. The more AGI-adjacent capability becomes embedded infrastructure across work, home and third place, the weaker the case becomes for treating it solely as a commercial convenience product.
- The field should resist allowing current product success to close down future pathways. The evidence here suggests that several of the most promising AGI-adjacent directions have been precisely those that current commercial dynamics under-resource. This includes work towards systems-level coordination, protected alignment research, open scientific collaboration, public-interest infrastructure and governance architectures that can survive under competitive pressures. The emergence of new ventures such as Thinking Machines may indicate not only technological ambition but an attempt to capture this next phase of the sector through yet another proprietary stack. This development makes the question of encouraging (and incentivising) plurality more urgent, not less. The strategic task for states, universities, citizenry and independent institutions is not simply to consume the next generation of capable models, but to ensure that this developmental pathway itself remains contestable.

A critical software studies perspective resists the fatalism that the current frontier-model pathway is the only future available. Software has always been contingent, historically layered and open to redesign even when dominant firms work hard to present their own arrangements as being inevitable. Alternatives are important for this reason. The significance of sovereign, public-interest and smaller-scale hybrid systems keep open the practical possibility of different software forms with different embedded values, institutional purposes and social contracts.

The overall conclusion of the report is not that commercialised generative AI was a mistake. It was the necessary bridge between lab capability and world-scale deployment. But it was a bridge with some specific directions already built into it. It accelerated one route by constraining so many others. If AGI-adjacent capability is to become something more than a profitable work-assistant embedded in a closed commercial stack, the next phase must widen out the field again. We need to move from attention to consumers to citizens, from the spectacle of benchmarks to intelligence within a viable system and from the competition of products to publicly governable systems capable of operating under complex ever changing conditions.

## 8 Appendices

- LLM Timeline – tracking key documents including model cards for the 5 key organisations and their products (excel document)
- Five (5) self-contained interactive HTML files documenting the Critical Pathway for each of the organisations.
- GEXF file documenting the key people relationships within and between these (and other) organisations
- Presentation based on this report (Powerpoint document)